%% file: main.tex
\documentclass{article}

\usepackage[utf8]{inputenc} 
\usepackage[T1]{fontenc}    
\usepackage{hyperref}       
\usepackage{url}            
\usepackage{booktabs}       
\usepackage{amsfonts}       
\usepackage{nicefrac}       
\usepackage{microtype}      
\usepackage{xcolor}         

\usepackage{graphicx} 
\usepackage{caption}
\usepackage{subcaption}
\usepackage{multirow}
\usepackage{chngpage}
\usepackage{tabularx}
\usepackage{amsmath}
\usepackage{nccmath}
\usepackage{comment}

\usepackage{paralist}
\usepackage{adjustbox}

\usepackage{wrapfig}
\usepackage{enumitem}
\usepackage[utf8]{inputenc}
\usepackage{tikz}
\usetikzlibrary{matrix,shapes,arrows,positioning,chains}

\newcommand*\samethanks[1][\value{footnote}]{\footnotemark[#1]}

\newcommand{\noop}[1]{{}}

\newcommand{\RK}[1]{{\color{red}{#1}}}

\title{Are all outliers alike? \\ A Taxonomy of Out of Distribution Samples}

\title{Are all outliers alike? \\ On Understanding the Diversity of Outliers\\ for Detecting OODs}

\usepackage[accepted]{icml2021}

\begin{document}

\twocolumn[
\icmltitle{Detecting OODs as datapoints with High Uncertainty}



\icmlsetsymbol{equal}{*}

\begin{icmlauthorlist}
\icmlauthor{Ramneet Kaur}{penn}
\icmlauthor{Susmit Jha}{sri}
\icmlauthor{Anirban Roy}{sri}
\icmlauthor{Sangdon Park}{penn}
\icmlauthor{Oleg Sokolsky}{penn}
\icmlauthor{Insup Lee}{penn}
\end{icmlauthorlist}

\icmlaffiliation{penn}{Department of Computer and Information Science, University of Pennsylvania, Philadelphia, USA.}
\icmlaffiliation{sri}{Computer Science Laboratory, SRI International, Menlo Park, CA 94025, USA}

\icmlcorrespondingauthor{Ramneet Kaur}{ramneetk@seas.upenn.edu}
\icmlcorrespondingauthor{Susmit Jha, Anirban Roy}{\{susmit.jha, anirban.roy\}@sri.com}
\icmlcorrespondingauthor{Sangdon Park, Oleg Sokolsky, Insup Lee}{\{sangdonp,sokolsky,lee\}@cis.upenn.edu}

\author{Ramneet Kaur\thanks{Ramneet Kaur is a graduate student at the Department of Computer and Information Science, University of Pennsylvania, Philadelphia, USA. This work was done when she was a summer intern at SRI.}\ , \; Susmit Jha\thanks{Computer Science Laboratory, SRI International, Menlo Park, CA 94025, USA} \ , Anirban Roy\samethanks\ , Sangdon Park\thanks{Department of Computer and Information Science, University of Pennsylvania, Philadelphia, USA.}, Oleg Sokolsky\samethanks, \ \& Insup Lee\samethanks  \\

\texttt{ramneetk@seas.upenn.edu,\{susmit.jha,anirban.roy\}@sri.com,} \\
\texttt{\{sangdonp, sokolsky, lee\}@cis.upenn.edu}
}

\icmlkeywords{Machine Learning, ICML}

\vskip 0.3in
]

\printAffiliationsAndNotice{\icmlEqualContribution}

\begin{abstract}
\input{abstract.tex}

\end{abstract}
\vspace{-0.8cm}
\input{intro.tex}

\vspace{-0.2cm}
\input{intuition.tex}
\vspace{-0.1cm}
\input{detection.tex}

\vspace{-0.25cm}
\input{exp.tex} 
\input{conclusion.tex}
\clearpage
\input{ack.tex}
\bibliography{biblio}
\bibliographystyle{icml2021}
\clearpage
\input{appendix.tex}

\end{document}

%% file: abstract.tex
Deep neural networks (DNNs) are known to produce incorrect predictions with very high confidence on out-of-distribution inputs (OODs). This limitation is one of the key challenges in the adoption of DNNs in high-assurance systems such as autonomous driving, air traffic management, and medical diagnosis. This challenge has received significant attention recently, and several techniques have been developed to detect inputs where the model's prediction cannot be trusted. These techniques detect OODs as datapoints with either high epistemic uncertainty or high aleatoric uncertainty. We demonstrate the difference in the detection ability of these techniques and propose an ensemble approach for detection of OODs as datapoints with high uncertainty (epistemic or aleatoric). We perform experiments on vision datasets with multiple DNN architectures, achieving state-of-the-art results in most cases.

%% file: intro.tex
\section{Introduction}
\label{sec:intro}
DNNs have achieved remarkable  performance in many areas such as computer vision~\citep{img-classification}, speech recognition~\citep{speech-recog}, and text analysis~\citep{text-analysis}. But their deployment in the safety-critical systems 
such as self-driving vehicles~\citep{ML-App7}, medical diagnoses~\citep{NNclinically} is hindered by their brittleness. 
One major challenge is the inability of DNNs to be self-aware of when new inputs are outside the training distribution and likely to produce incorrect predictions. It has been widely reported in literature~\citep{guo2017calibration,baseline} that DNNs 
exhibit overconfident incorrect predictions on inputs which are outside the training distribution. 
The responsible deployment of DNNs in high-assurance applications necessitates detection of out-of-distribution datapoints (OODs) so that DNNs can abstain from making decisions on those. 


OODs are those points that do not belong to any class of the in-distribution (iD). Existing techniques detect OODs as datapoints that have either lack of support from the iD data~\citep{mahalanobis, vae-recon-err} or high entropy in the class prediction~\citep{baseline, kl_div_entropy}. Lack of support from the iD data indicates high epistemic uncertainty (EU) and high entropy in the class prediction indicates high aleatoric uncertainty~\citep{ML-uncertainty-survey}. Existing techniques can thus be classified into two categories; one that detect OODs due to high EU and other that detect OODs due to high AU. We propose detecting OODs as datapoints with high uncertainty (aleatoric or epistemic).

High entropy in the predictive distribution by the 
\textit{ensemble of classifiers} has been proposed by~\citet{proper_scoring} for OOD detection. We propose OOD detection with an ensemble of detectors where each detector is composed of indicators for both, high EU and high AU.

We make the following three contributions in this paper: 
\begin{compactitem}
    
    \item \textbf{Classification of OOD detection techniques.}  We classify the existing techniques as detecting OODs due to either high EU or high AU.
    
    \item \textbf{OODs as datapoints with high uncertainty.} We illustrate difference in the detection abilities of the classified techniques and propose detecting OODs as datapoints with high uncertainty (aleatoric or epistemic).
    
    \item \textbf{Ensemble approach to detect OODs.} We propose a novel OOD detection technique based on the ensemble of OOD detectors where each detector is composed of indicators for both, high EU and high AU.
    
    \item \textbf{Empirical evaluation.} We demonstrate the effectiveness of our approach on several vision benchmarks, obtaining state-of-the-art (SOTA) results.
    
\end{compactitem}

\noop{

\section{Introduction}
\label{sec:intro}
Deep neural networks (DNNs) have achieved near human-level accuracy in many domains 
such as computer vision~\citep{img-classification}, speech recognition~\citep{speech-recog}, and text analysis~\citep{text-analysis}. But their deployment in the safety-critical systems 
such as self-driving vehicles~\citep{ML-App7}, aircraft collision avoidance~\citep{NNinAircraft}, and medical diagnoses~\citep{NNclinically} is hindered by their brittleness and the resulting lack of trust.
One major challenge is the inability of DNNs to be self-aware of when these models are outside their training distribution and likely to produce incorrect predictions. It has been widely reported in literature~\cite{guo2017calibration,baseline} that deep neural networks overfit in the negative log likelihood space and consequently exhibit overconfident predictions even on inputs which are from a different distribution far from the training data and likely to produce wrong predictions. The responsible deployment of deep neural network models in high assurance applications necessitates detection of  out-of-distribution (OOD) data so that DNNs can abstain from making decisions on those. }

%% file: intuition.tex
\section{Background}
\label{sec:intuition}
OOD datapoints (OODs) do not belong to (any class of) the in-distribution (iD). So, OODs can be detected by:

\setlength{\parindent}{0em}

\begin{enumerate}[nosep,leftmargin=1em,labelwidth=*,align=left]
    \item Lack of support (or evidence) from the iD data to make decision on these points as they do not belong to the iD; indicating high \textit{epistemic uncertainty (EU)}~\citep{ML-uncertainty-survey}.
    \item High entropy in the class prediction as they do not belong to any iD class; indicating high \textit{aleatoric uncertainty (AU)}~\citep{ML-uncertainty-survey}.
\end{enumerate}

\begin{figure}[!t]
\begin{subfigure}{0.25\textwidth}
  \includegraphics[width=0.8\linewidth]{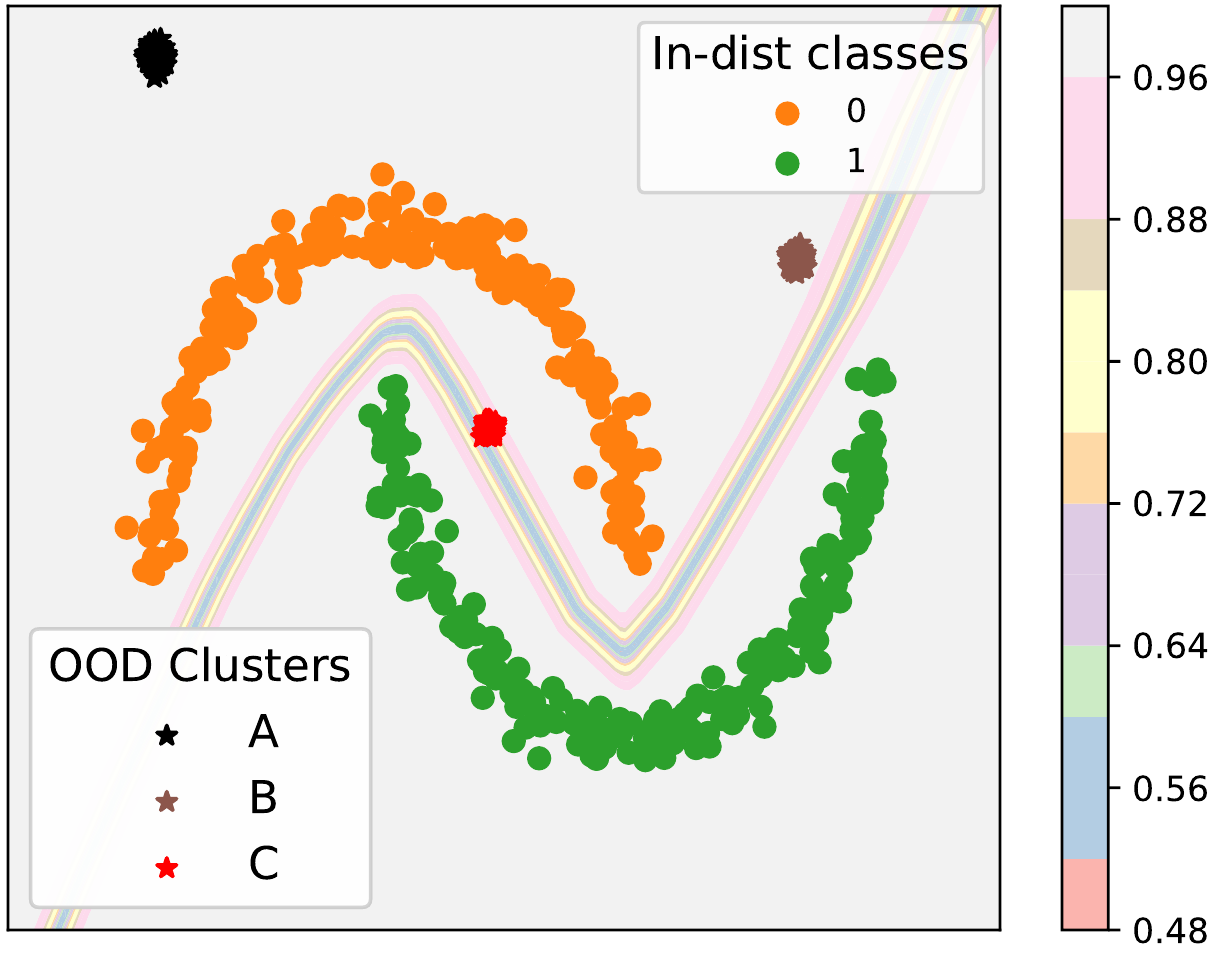}
\end{subfigure}%
\begin{subfigure}{0.25\textwidth}
  \includegraphics[width=0.8\linewidth]{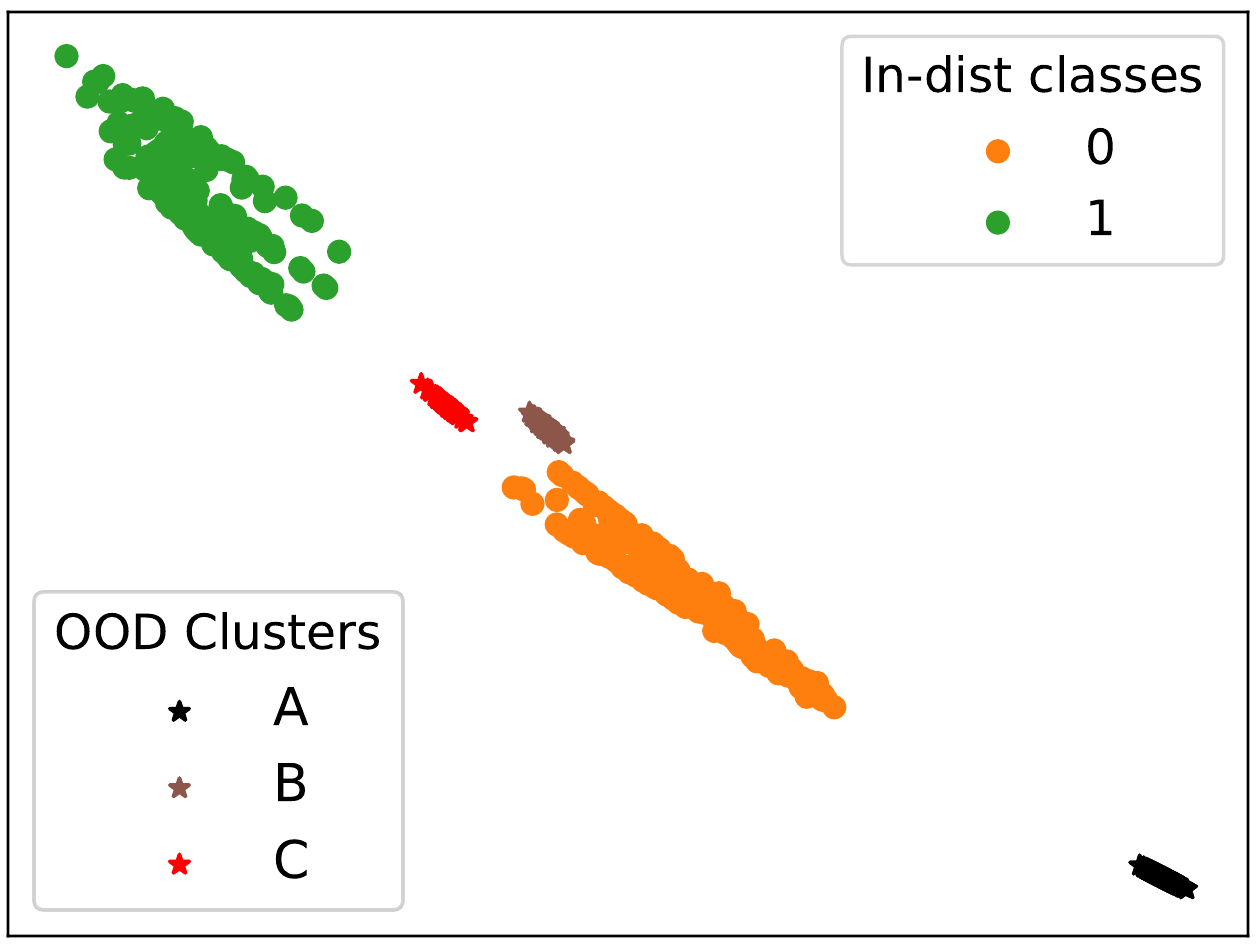}
\end{subfigure}%
\caption{
(Left) shows the 2 iD half-moon classes, 3 OOD clusters, and the trained classifier's boundary with softmax scores in the input space. (Right) shows iD samples and OODs in the classifier's penultimate feature space.} 
\vspace{-0.25in}
\label{fig:toy_ex}
\end{figure}

\begin{figure*}[!h]
    \centering
    \subcaptionbox{Mahalanobis}
      {\includegraphics[width=.2\linewidth]{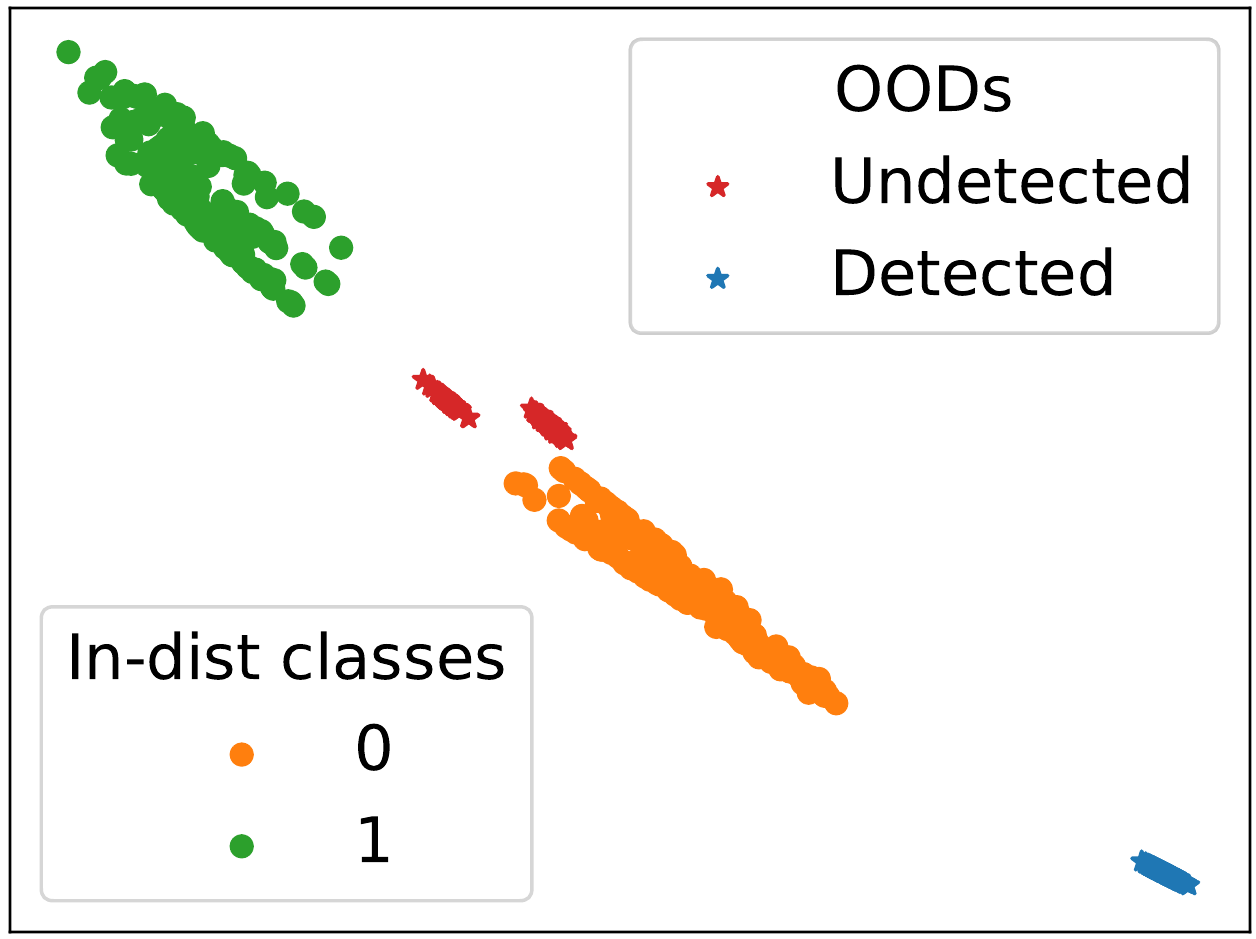}}
  \quad
    \centering
    \subcaptionbox{PCA}
      {\includegraphics[width=.2\linewidth]{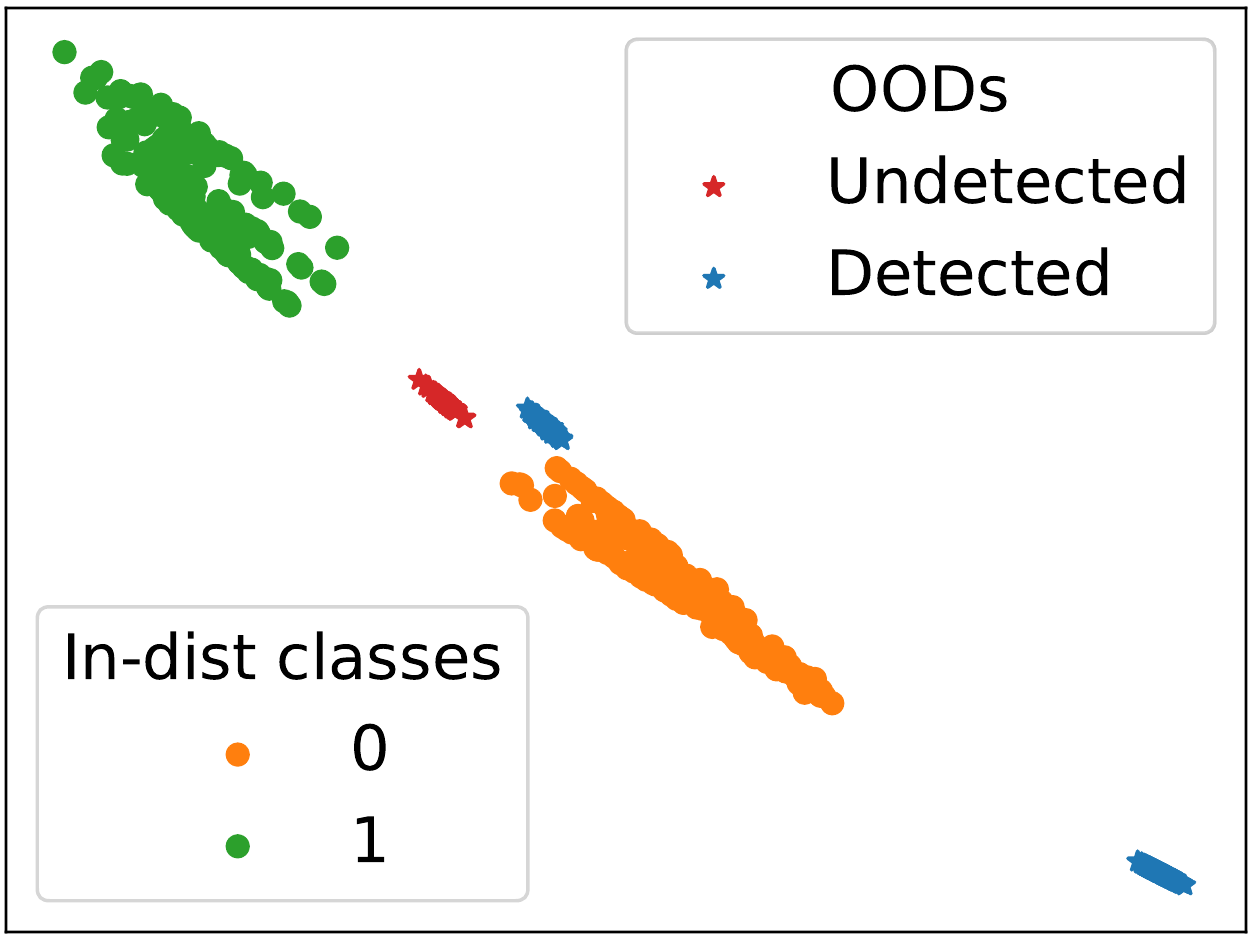}}
\quad
    \centering
    \subcaptionbox{SBP}
      {\includegraphics[width=.2\linewidth]{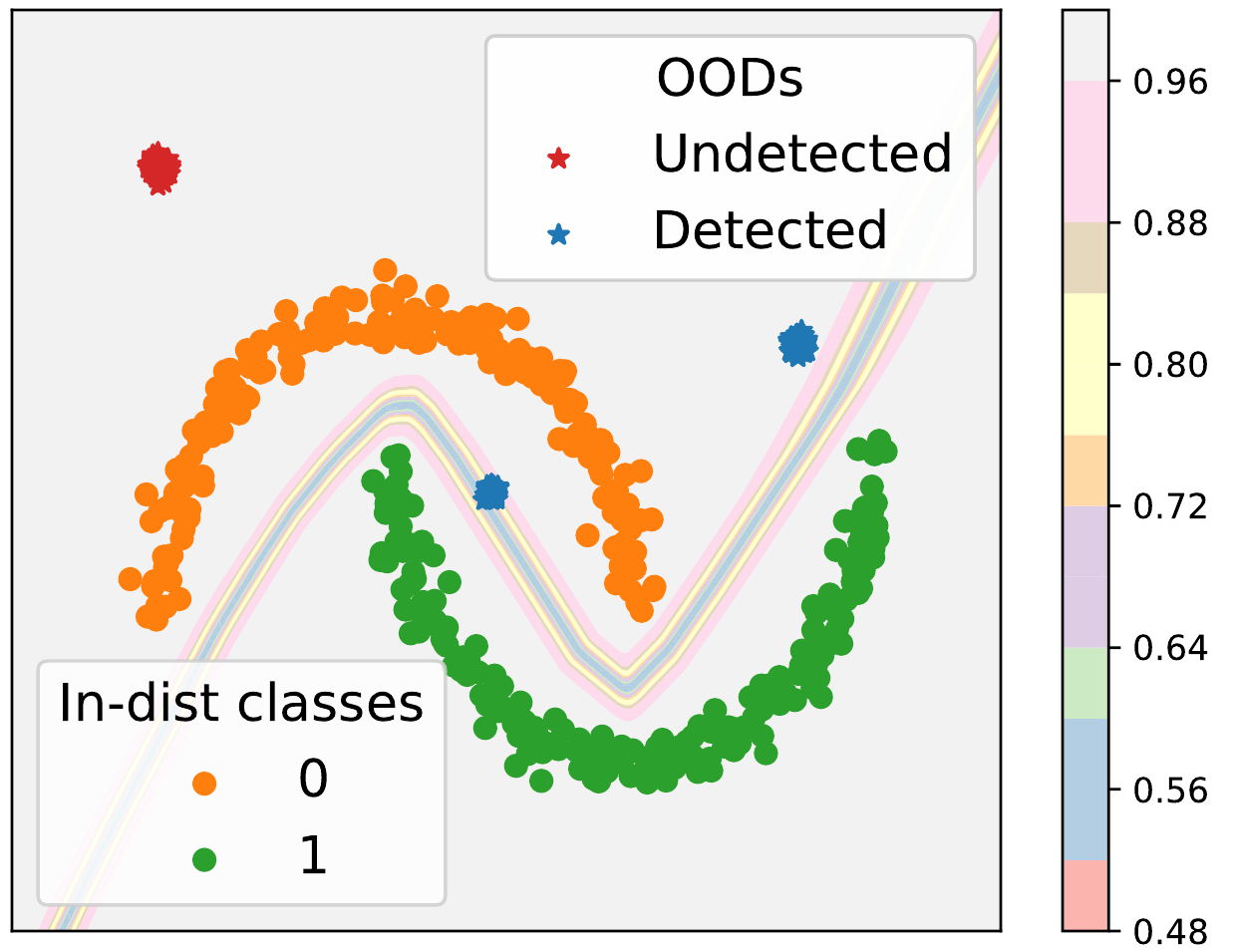}}
  \quad
    \centering
    \subcaptionbox{DkNN}
      {\includegraphics[width=.2\linewidth]{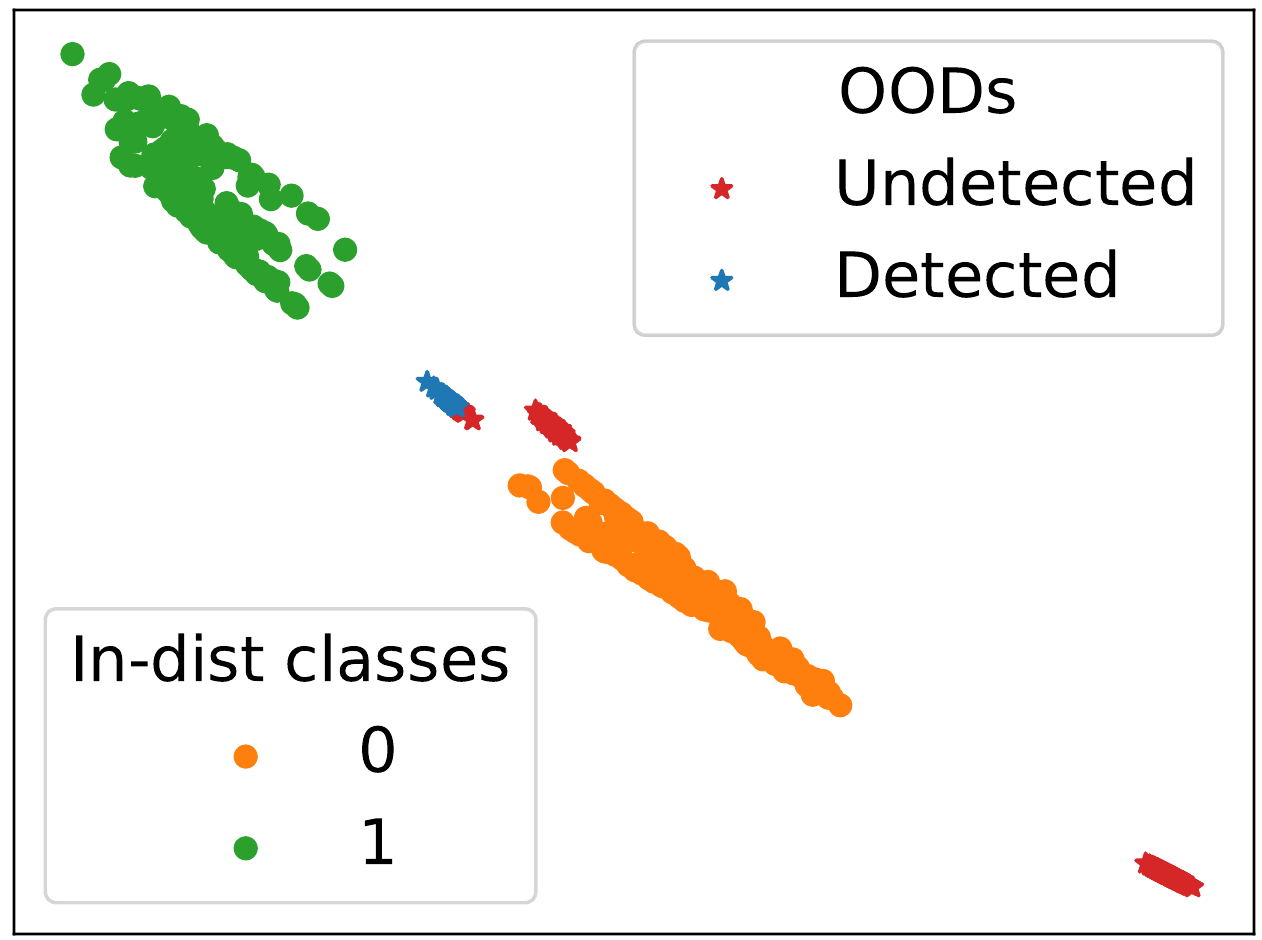}}
\caption{\textit{Different techniques differ in their ability to detect OODs.}
}
\label{fig:failure_toy_ex}
\vspace{-0.1in}
\end{figure*}

\textbf{Existing OOD detection techniques.}
Current approaches for OOD detection detect OODs as datapoints either with high EU or high AU. Therefore, they differ in their ability to detect OODs. 
We demonstrate this difference on a 2D half-moon dataset. As shown in Figure~\ref{fig:toy_ex}, we consider three clusters of OODs: cluster $A$ (black), $B$ (brown) and $C$ (red).
We consider two approaches for detecting OODs with EU and two approaches for detecting OODs with high AU:

\begin{itemize}[nosep,leftmargin=1em,labelwidth=*,align=left]

    \item \citet{mahalanobis} propose using Mahalanobis distance of an input from the iD density for OOD detection. This corresponds to using lack of support from the iD data, i.e., large distance from the iD density to detect OODs as datapoints with high EU. Figure~\ref{fig:failure_toy_ex}(a) shows that the Mahalanobis distance from the mean and covariance of all the iD data in the penultimate feature space is only able to detect the cluster $A$ that lie far from the iD density. 
    
    \item Using reconstruction error from the Principal Component Analysis (PCA)~\citep{pca} on the iD data is another approach that detects OODs as data points with high EU. This is because it uses lack of support from the iD data, i.e., high reconstruction error from the PCA of the iD data for OOD detection. Figure~\ref{fig:failure_toy_ex}(b) shows that using minimum reconstruction error from the class-conditional PCA performed in the feature space of the iD data is able to detect OODs from clusters $A$ and $B$.  
    

    \item \citet{baseline} propose using maximum softmax probability as an indicator of the KL-divergence between distribution of the predicted softmax probabilities and the uniform distribution. This divergence measures entropy in the prediction of the class for an input~\citep{kl_div_entropy} and therefore detects OODs as data points with high AU. Figure~\ref{fig:failure_toy_ex}(c) shows that this technique (SBP) is able to detect those OODs that lie on or near the decision boundary where the classifier is least confident or has high entropy in its prediction. 
     
    \item Another approach for detection of OODs as data points with high AU is by using non-conformance in the labels of the K-Nearest Neighbors for OOD detection (DkNN)~\citep{dknn}. As shown in Figure~\ref{fig:failure_toy_ex}(d), entropy in the label of the kNNs from the penultimate layer is only able to detect OODs with nearest neighbors from multiple classes (cluster $C$). 

\end{itemize}

Existing techniques for detecting OODs due to either high AU or high EU are summarized in Table~\ref{tab:detect_oods_eu_au} of appendix.

\begin{figure}
    \centering
      {\includegraphics[width=.3\linewidth]{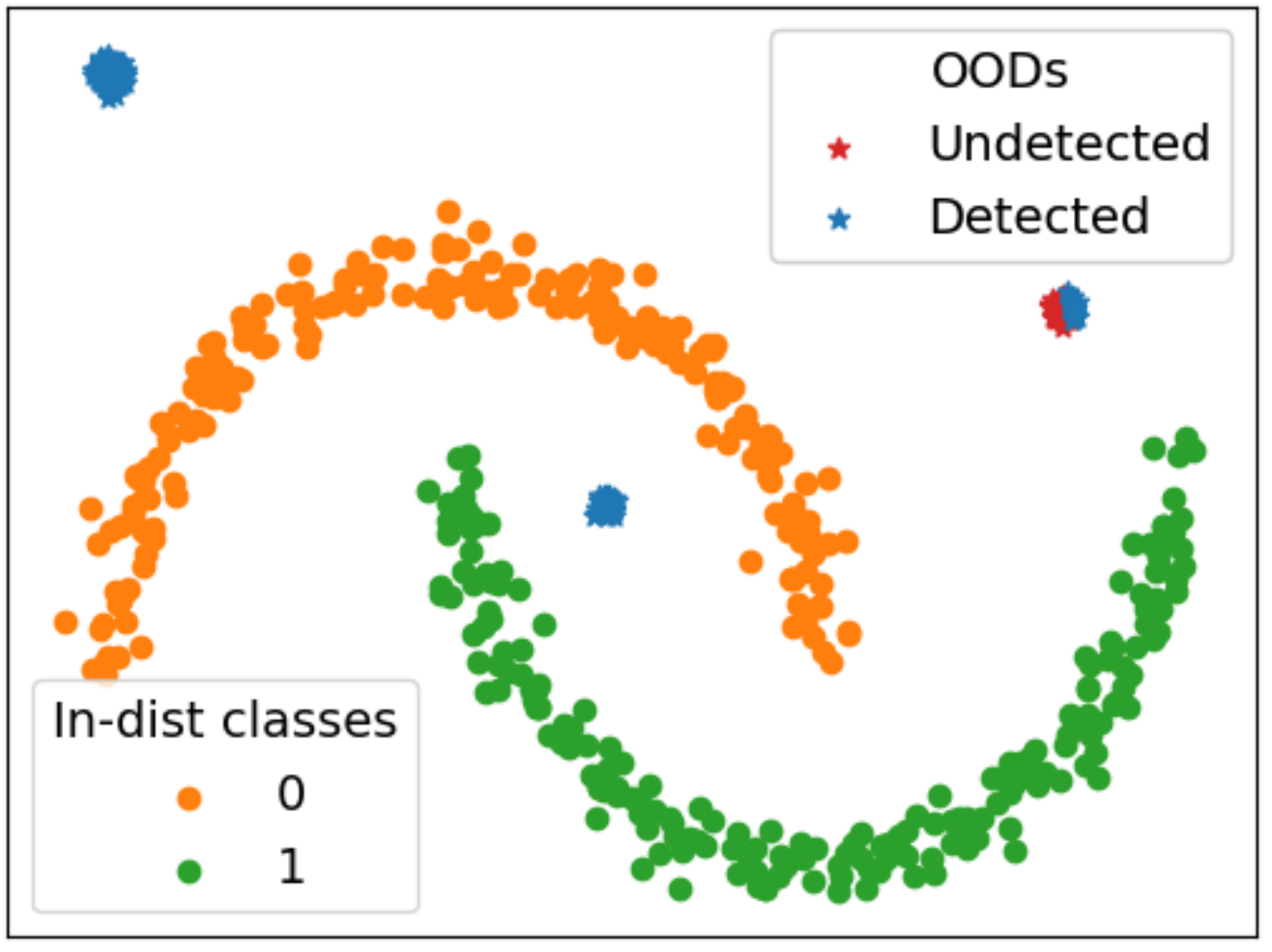}}
  \quad
    \centering
      {\includegraphics[width=.3\linewidth]{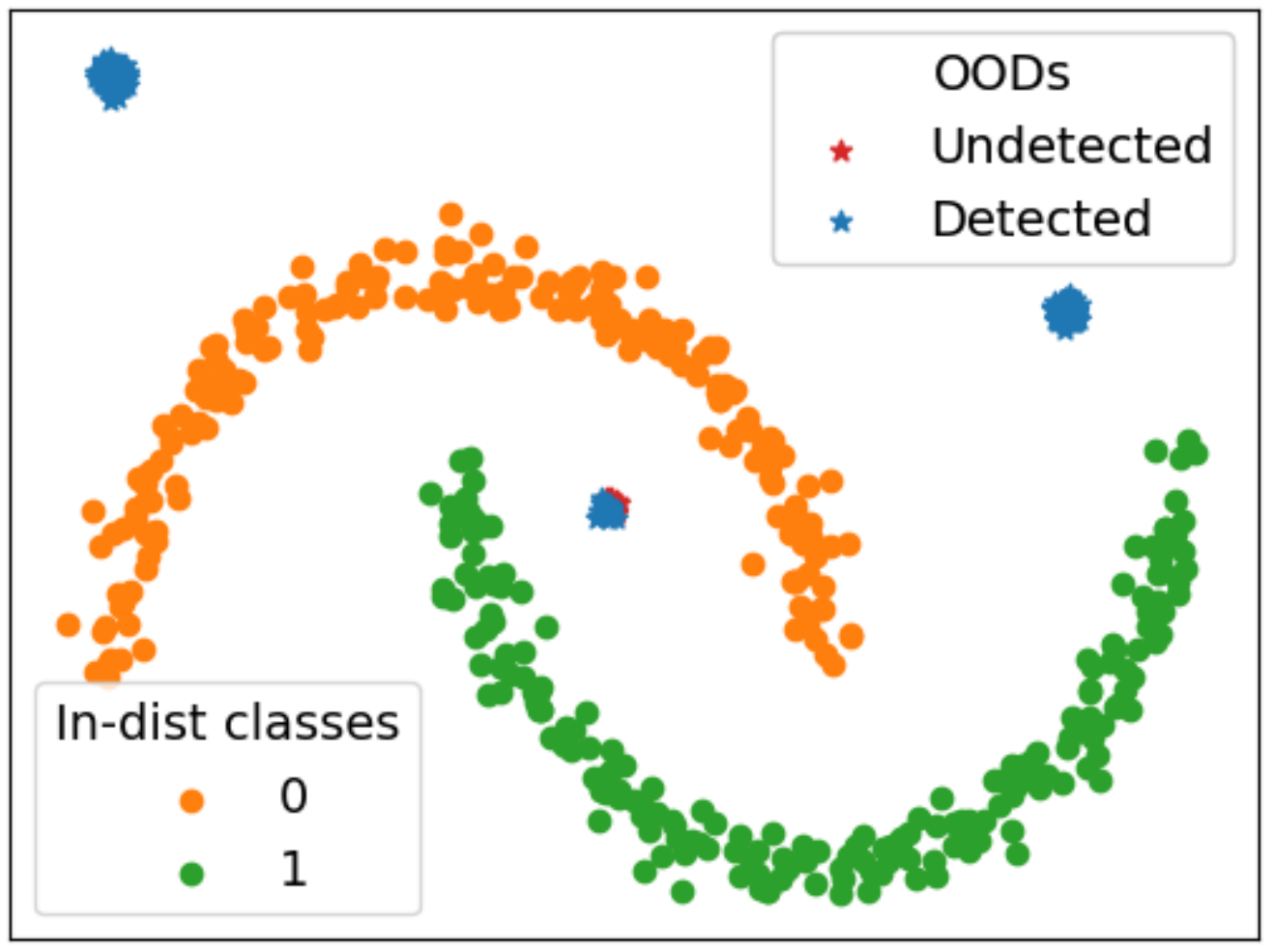}}
    \centering
      {\includegraphics[width=.3\linewidth]{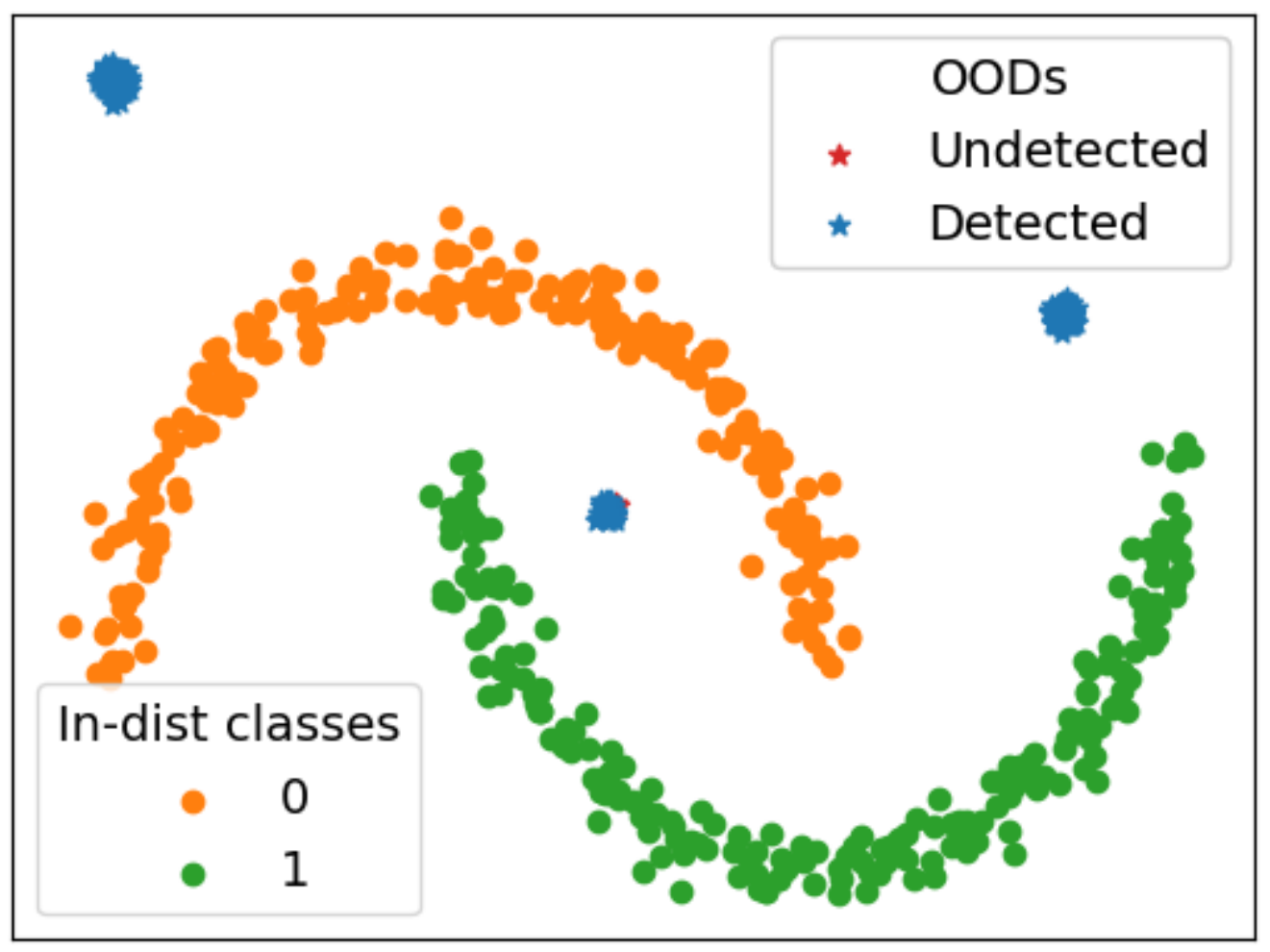}}

\caption{Ensemble of OOD detectors improves TNR at 95\% TPR. (Left) SBP (for AU) and Mahalanobis (for EU) detects \textbf{62.73\%} OODs. (Middle) DkNN (for AU) and PCA (for EU) detects \textbf{95.91\%} OODs. (Right) Ensemble of both detectors (from left and right) detects \textbf{99.55\%} OODs.}
\label{fig:MoE_intuition}
\vspace{-0.15in}
\end{figure}

\textbf{Proposed approach for OOD detection.} 
We propose using both lack of support from the iD data as well as high entropy in the class prediction to detect OODs as datapoints with high uncertainty (epistemic or aleatoric). Further, as observed from the toy example, techniques for detecting OODs as datapoints with high EU (or AU) such as Mahalanobis and PCA (or SBP and DkNN) also differ in their abilities. Therefore, we propose using an ensemble of OOD detectors where each detector is composed of indicators for both, high EU and high AU. Figure~\ref{fig:MoE_intuition} shows improvement in the True Negative Rate (TNR) of the proposed technique with two detectors over a single detector on the two half-moons dataset.\footnote{Empirical evaluation on CIFAR10 and SVHN in appendix~\ref{apdx:ablation} also justify these observations. We compare the performance of ensemble approach with the indicators of high AU and high EU as well as individual detectors used in the ensemble approach. Our approach achieves SOTA in almost all cases.} 




%% file: detection.tex
\section{Ensemble approach for detection of OODs as datapoints with high uncertainty}
\label{sec:detection}
\textbf{OODs as datapoints with high AU.}
For a given threshold $\delta_a$ on the entropy in the predicted class distribution for the iD data and $n_c$ as the number of classes, an input $x$ is detected as an OOD due to high AU if $-\sum\limits_{i=1}^{n_c} p_{i|x} \ \log\ (p_{i|x}) > \delta_a.$
Here $p_{i|x}$ is the predicted probability of $x$ in class $i$.

\textbf{OODs as datapoints with high EU.}
Probability density function (PDF) estimated from the iD data can be used to provide support for a datapoint belonging to the iD. For a given threshold $\delta_e$ on the probability of an input $x$ belonging to the iD density and $\{q_i(.)\ : \ 1 \leq i \leq n_c \}$ as the class-conditional iD PDF set, $x$ is detected as an OOD due to high EU if
$\displaystyle \max_{i=1}^{n_c}\{q_i(x)\} < \delta_e.$


\noop{Further, typical inputs to
DNNs are high-dimensional and can be decomposed into principal and non-principal components based on the direction of high variation. So, one can consider only principal components in the PDF of the iD data to detect OODs due to
high deviation in the principal components of the iD data. With $q_{p}$ as the PDF with only principal components of the iD data and $\delta_{ep}$ as the threshold on the probability of an input belonging to the iD density with the principal components, $x$ is detected as an OOD due to high epistemic uncertainty if:
\begin{equation}
    q_{p}(x) < \delta_{ep}
\label{eq_p_np_tied}
\end{equation}

$q(x)$ defined so far is for the tied (or single) iD. For class-conditional iD density, we will have a set $\{q_i(x)\ | \ 1 \leq i \leq n_c \}$ for PDFs for each class and thus the following conditions for OOD detection. 

With $\{q_i(x)\ | \ 1 \leq i \leq n_c \}$ as the class-conditional PDF set for the iD, an input $x$ is detected as an OOD with high epistemic uncertainty if:

\begin{equation}
    \max_{i=1}^{n_c}\{q_i(x)\} < \delta_e
\label{eq_cc}
\end{equation}

Similarly, considering only principal $\{q_{pi}(x)\ | \ 1 \leq i \leq n_c \}$  components in defining the  class-conditional iD density detects an input $x$ as OOD with high epistemic uncertainty if:

\begin{equation}
    \max_{i=1}^{n_c}\{q_{pi}(x)\} < \delta_{ep} 
\label{eq_p_np_cc}
\end{equation}

Equations~\ref{eq_tied} and~\ref{eq_p_np_tied} are the special case of Equations~\ref{eq_cc} and~\ref{eq_p_np_cc} respectively where the size of the sets representing the PDF of iD is equal to 1. Also, Equations~\ref{eq_tied} and~\ref{eq_cc} are a special case of Equations~\ref{eq_p_np_tied} and~\ref{eq_p_np_cc} respectively, when we do not consider dimensionality reduction, i.e. all components are considered as principal components in the PDF of iD.}

\textbf{OODs as data points with high uncertainty.} We detect an input $x$ as an OOD if it has high AU or high EU:
\begin{equation}
 -\sum\limits_{i=1}^{n_c} p_{i|x} \ \log\ (p_{i|x}) > \delta_a \vee \displaystyle \max_{i=1}^{n_c}\{q_{i}(x)\} < \delta_{e}.
 \label{eq:ood_due_to_uncertainty}
\end{equation}
There are different ways of assigning score to the OOD nature of an input $x$ from~\eqref{eq:ood_due_to_uncertainty}. We call these scores as uncertainty scores. One way to compute the uncertainty score is by picking the dominating uncertainty:
\begin{align}
     \max (-\sum_{i=1}^{n_c} p_{i|x} \ \log\ (p_{i|x}) - \delta_a, \delta_{e} - \max_{i=1}^{n_c}\{q_{i}(x)\})
\label{eq:max_ood_score}
\end{align}
Another way is to use the linear combination of AU and EU:
\begin{align}
\label{eq:lin_ood_score}
     w_1\times(-\sum\limits_{i=1}^{n_c} p_{i|x} \ \log\ (p_{i|x}))  \\ \nonumber + \ 
     w_2 \times(\displaystyle \max_{i=1}^{n_c}\{q_{i}(x)\})
\end{align}
There can be other ways of assigning uncertainty score to the input. We use the score from~\eqref{eq:lin_ood_score} in our experiments. 

\textbf{Ensemble approach for OOD detection.} We propose OOD detection by combining uncertainty scores by different detectors. There are multiple ways of assigning weights to the predictions of individual detectors in an ensemble approach~\citep{ensemble}. We use logistic regression for assigning weights to the uncertainty scores by individual detectors in our experiments.

\noop{One can also consider only principal (or non-principal) components in the PDF of the iD data to detect OODs due to high (or smaller) deviation in the principal (or non-principal) components of the iD. With $q_{p}$ (or $q_{n}$) as the PDF with only principal (or non-principal) components of the iD data and $\delta_{ep}$ (or $\delta_{en}$) as the threshold on the probability of an input belonging to the iD density due to the principal (or non-principal) components, $x$ is detected as an OOD due to high epistemic uncertainty if:
\begin{equation}
    q_{p}(x) < \delta_{ep} \vee q_{n}(x) < \delta_{en}\  \textrm{where} \  \delta_{ge} > \delta_{ge}
\label{eq_p_np_tied}
\end{equation}
\label{lem:tied_p_np}
While considering only principal (or non-principal) components in the PDF of iD, the other term in the above equation corresponding to the non-principal (or principal) components is considered false. In other words, an input is assumed to be iD along those components for which it is not tested. However, if analysis is performed for both the components separately (instead of together as in Equation~\ref{eq_tied}), then the input is detected as an OOD if either of the two conditions in Equation~\ref{lem:tied_p_np} is true.

Note that $q(x)$ defined so far is for the tied (or single) iD. For class-conditional iD density, we will have a set $\{q_i(x)\ | \ 1 \leq i \leq n_c \}$ for PDFs for each class and thus the following conditions for OOD detection. 

With $\{q_i(x)\ | \ 1 \leq i \leq n_c \}$ as the class-conditional PDF set for the iD, an input $x$ is detected as an OOD if:

\begin{equation}
    \max_{i=1}^{n_c}\{q_i(x)\} < \delta_e
\label{eq_cc}
\end{equation}
\label{lem:cc}

Similarly, considering only principal $\{q_{pi}(x)\ | \ 1 \leq i \leq n_c \}$ (or non-principal $\{q_{ni}(x)\ | \ 1 \leq i \leq n_c \}$) components in defining the  class-conditional iD density detects an input $x$ as OOD if:

\begin{equation}
    \max_{i=1}^{n_c}\{q_{pi}(x)\} < \delta_{ep} \vee \max_{i=1}^{n_c}\{q_{ni}(x)\} < \delta_{en} \ \textrm{where} \  \delta_{ep} > \delta_{en}
\label{eq_p_np_cc}
\end{equation}
\label{lem:cc_p_np}

Note that Equations~\ref{eq_tied} and~\ref{eq_p_np_tied} are the special case of Equations~\ref{eq_cc} and~\ref{eq_p_np_cc} respectively where the size of the sets representing the PDF of iD is equal to 1. Also, Equations~\ref{eq_tied} and~\ref{eq_cc} are a special case of Equations~\ref{eq_p_np_tied} and~\ref{eq_p_np_cc} respectively, when we do not consider dimensionality reduction, i.e. all components are considered as principal components in the PDF of iD. 
}

%% file: exp.tex
\section{Experimental Results}
\label{sec:exp}
\textbf{Individual detectors. } We use two detectors in the proposed ensemble approach for OOD detection. First detector is composed of the Mahalanobis distance (for EU)\footnote{Mahalanobis distance of an input from the estimated iD density corresponds to measuring the log of probability densities of an input~\citep{mahalanobis}. Using mahalanobis distance from the empirical class means and tied empirical covariance of all the training iD data thus detects OODs as datapoints with high EU.} and ODIN~\citep{odin} with $\epsilon = 0.005$ and $T=10$ (for AU). Second detector is composed of the minimum reconstruction error from the class-conditional PCA on the top 40\% eigen vectors (for EU)\footnote{PCA can be viewed as a maximum likelihood procedure on a Gaussian density model of the observed data~\citep{prob_pca}. Performing class-wise reconstruction error from PCA thus detects OODs as datapoints with high EU.} and a novel non-conformance measure amongst nearest neighbors\footnote{Details are given in the appendix.} (for AU).

\textbf{Evaluation. } We evaluate the proposed technique on different iD datasets on different architectures. We consider MNIST~\citep{lenet}, CIFAR10~\cite{cifar10} and SVHN~\citep{svhn} as iD datasets. KMNIST~\citep{kmnist} and F-MNIST~\cite{fashion-mnist} datasets are considered as OOD for MNIST. For CIFAR10 and SVHN, we consider LSUN~\citep{lsun}, Imagenet~\citep{imagenet}, SVHN (for CIFAR10 as iD) and CIFAR10 (for SVHN as iD) as OOD. We also consider a Subset-CIFAR100 as OODs for CIFAR10 and SVHN. Specifically, from the CIFAR100 classes, we select sea, road, bee, and butterfly as OOD which are visually similar (and thus challenging OOD for CIFAR10) to the ship, automobile, and bird classes in the CIFAR10, respectively. We report TNR at 95\% TPR, area under receiver operating characteristic curve (AUROC), and detection accuracy (DTACC). We compare with the SOTA detectors that are used as indicators of AU or EU in our approach; namely SBP~\citep{baseline}, ODIN~\citep{odin} and Mahalanobis~\citep{mahalanobis}.

\begin{figure}[!h]
  \centering
  \includegraphics[width=1\linewidth]{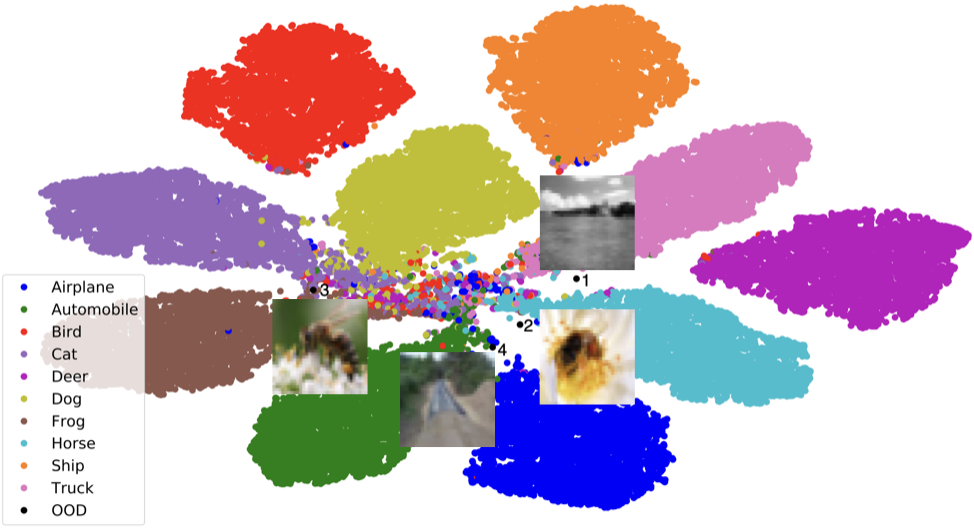}
\caption{t-SNE plot of the penultimate layer feature space of ResNet50 trained on CIFAR10. We show four OOD images from the SCIFAR100.
OOD 1 and OOD 2 are far from the distributions of all classes and thus represent OODs due to high EU. OOD 3 and OOD 4 are OODs due to high AU as they lie closer to two class distributions. Third OOD is closer to the cat and frog classes of the ID and forth OOD is closer to the airplane and automobile classes of the ID.}
\vspace{-0.3in}
\label{fig:mahala_our_tsne}
\end{figure}

\textbf{Results. } Table~\ref{main_table} shows that the proposed ensemble approach for detecting OODs as datapoints with high uncertainty outperforms SOTA in almost all the cases. Figure~\ref{fig:mahala_our_tsne} shows the t-SNE~\citep{t-SNE} plot of the penultimate features from the ResNet50 model trained on CIFAR10.  We show 4 examples of OODs (2 due to high EU and 2 due to high AU) from Subset-CIFAR100. These OODs were detected by the proposed approach but missed by the Mahalanobis approach. 

\begin{table*}[!t]
\caption {\label{tab:comp_base_icad}\footnotesize{Comparison with SBP, ODIN and Mahalanobis as SOTA OOD detection techniques.}} 
\begin{adjustbox}{width=1.75\columnwidth,center}
\begin{tabular}{c|cccc|cccc|cccc}
\hline
MNIST (LeNet5) & \multicolumn{4}{c|}{TNR (95\% TPR)}  &   \multicolumn{4}{c|}{AUROC}  &   \multicolumn{4}{c}{DTACC}   \\ 
\cline{1-5} \cline{6-8}  \cline{9-13}
$D_{out}$  & SBP      & ODIN     & Mahala  & Ours                                                         
	&  SBP & ODIN    & Mahala   & Ours    & SBP & ODIN     & Mahala  & Ours                                          \\ 
\hline
KMNIST  & 69.33 & 67.72  & 80.52   & \textbf{91.7} 
 &  93.24 & 92.98 & 96.53  & \textbf{98.29} 
& 86.88	& 85.99 &  90.82 &\textbf{93.98}\\
	 			
F-MNIST & 52.69  &  58.47  &   63.33   & \textbf{72.62}
& 89.19   & 90.76  & 94.11  & \textbf{95.49}
& 82.77	&  83.21 & 87.76 &\textbf{90.56}  \\
		
\hline

\hline
CIFAR-10 (ResNet34) & \multicolumn{4}{c|}{TNR (95\% TPR)}  &   \multicolumn{4}{c|}{AUROC}  &   \multicolumn{4}{c}{DTACC}   \\ 
\cline{1-5} \cline{6-8}  \cline{9-13}
$D_{out}$       & SBP      & ODIN     & Mahala  & Ours                                                         
	&  SBP & ODIN    & Mahala   & Ours    & SBP & ODIN     & Mahala  & Ours                                          \\ 

\hline
SVHN  & 32.47 & 72.85   & 53.16  & \textbf{83.2} 
&  89.88  & 93.85 & 93.85  & \textbf{96.91} 
& 85.06	& 85.40 & 89.17 &\textbf{91.16}\\
LSUN & 45.44  & 45.16  &   77.53  & \textbf{81.23}
& 91.04   & 89.63  & 96.51 & \textbf{96.87}
& 85.26	& 81.83  & 90.64 &\textbf{91.19}  \\
ImageNet & 44.72 &  46.54  &  68.41  & \textbf{74.53}
&  91.02  & 90.45 & 95.02 & \textbf{95.73}
& 85.05	& 83.06 & 88.63 &	\textbf{89.73}\\
SCIFAR100 &38.17 & 37.00   & 38.39  &   \textbf{51.11}
 & 88.91  & 86.13  & 88.86 & \textbf{93.85}
& 82.34	& 78.50 & 82.51 &	\textbf{89.93}\\
\hline

\hline
CIFAR-10 (ResNet50) & \multicolumn{4}{c|}{TNR (95\% TPR)}  &   \multicolumn{4}{c|}{AUROC}  &   \multicolumn{4}{c}{DTACC}   \\ 
\cline{1-5} \cline{6-8}  \cline{9-13}
$D_{out}$        & SBP      & ODIN     & Mahala  & Ours                                                         
	&  SBP & ODIN    & Mahala   & Ours    & SBP & ODIN     & Mahala  & Ours                                          \\ 
\hline
SVHN  & 44.69 & 86.61   & 34.49   & \textbf{88.8} 
 & 97.31  & 84.41 & \textbf{98.19}  & 97.84
& 86.36	& 91.25 & 76.72 &\textbf{92.26}\\
	 			
LSUN & 48.37 & 80.72  &   32.18   & \textbf{81.38}
&  92.78  & 96.51  & 87.09  & \textbf{96.93}
& 86.97	&  90.59 & 80.07 &\textbf{91.79}  \\
			
ImageNet & 42.06  & 73.23   &  29.48  & \textbf{74.44}
 & 90.80  & 94.91 & 84.30 & \textbf{95.6}
& 84.36	& 88.23 & 77.19 &	\textbf{89.42}\\
			
SCIFAR100 & 36.39  & 47.44   & 21.06   &   \textbf{48.33}
&  89.09 & 86.16  & 77.42 & \textbf{92.98}
& 83.37	& 78.69 & 71.43 &	\textbf{88.27}\\
			
\hline

\hline
CIFAR-10 (DenseNet) & \multicolumn{4}{c|}{TNR (95\% TPR)}  &   \multicolumn{4}{c|}{AUROC}  &   \multicolumn{4}{c}{DTACC}   \\ 
\cline{1-5} \cline{6-8}  \cline{9-13}
$D_{out}$        & SBP      & ODIN     & Mahala  & Ours                                                         
	&  SBP & ODIN    & Mahala   & Ours    & SBP & ODIN     & Mahala  & Ours                                          \\  
\hline
SVHN  & 39.22 & 69.96   & 83.63   & \textbf{90.92} 
 &  88.24 & 92.02 & 97.10  & \textbf{98.41} 
& 82.41	& 84.10 & 91.26 &\textbf{93.29}\\
	 					
LSUN & 48.38 & 71.89  &   46.63   & \textbf{83.47}
&  92.14  & 94.37 & 91.18  & \textbf{97.07}
& 86.22	&  87.72 & 84.93  &\textbf{91.74}  \\
					
ImageNet & 40.13 & 61.03   &  49.33  & \textbf{77.56}
& 89.30   & 91.40 & 90.32 & \textbf{95.86}
& 82.67	& 83.85 & 83.08 &	\textbf{89.55}\\
					
SCIFAR100 & 34.11 & \textbf{35.06}   & 20.33   &   32.11
&  85.53  & 80.18  & 80.40 & \textbf{90.09}
& 79.18	& 72.58 & 74.15 &	\textbf{85.2}\\
\hline

\hline
SVHN (ResNet34) & \multicolumn{4}{c|}{TNR (95\% TPR)}  &   \multicolumn{4}{c|}{AUROC}  &   \multicolumn{4}{c}{DTACC}   \\ 
\cline{1-5} \cline{6-8}  \cline{9-13}
$D_{out}$        & SBP      & ODIN     & Mahala  & Ours                                                         
	&  SBP & ODIN    & Mahala   & Ours    & SBP & ODIN     & Mahala  & Ours                                          \\ 
\hline
CIFAR10  & 78.26 & 32.60   & 85.03   & \textbf{90.34} 
&  92.92  & 66.75 & 97.05  & \textbf{97.64} 
& 90.03	&65.37 & 93.15 &\textbf{94.29}\\
	 							
LSUN & 74.29  & 35.92  &   78.38   & \textbf{85.46}
&  91.58  & 68.60 & 96.17  & \textbf{97.09}
& 88.96	& 66.75  & 91.98 &\textbf{93.17}  \\
							
ImageNet & 79.02 &  41.80   &  84.46  & \textbf{89.81}
&  93.51  & 73.00 & 96.95 & \textbf{97.6}
& 90.44	& 60.84 & 93.14 &	\textbf{94.32}\\
							
SCIFAR100 & 81.28 & 36.67   & 86.61   &   \textbf{97.28}
& 94.62   & 68.01  & 97.30 & \textbf{98.19}
& 91.48	& 67.26 & 93.60 &	\textbf{96.39}\\
			
\hline

\hline
SVHN (DenseNet) & \multicolumn{4}{c|}{TNR (95\% TPR)}  &   \multicolumn{4}{c|}{AUROC}  &   \multicolumn{4}{c}{DTACC}   \\ 
\cline{1-5} \cline{6-8}  \cline{9-13}
$D_{out}$        & SBP      & ODIN     & Mahala  & Ours                                                         
	&  SBP & ODIN    & Mahala   & Ours    & SBP & ODIN     & Mahala  & Ours                                          \\ 
\hline
CIFAR10 & 69.31 &  37.23   & 80.82   & \textbf{83.81} 
    & 91.90  & 73.14 & 96.80  & \textbf{97.17} 
	& 86.61 & 68.92 & \textbf{92.87} & 92.69\\
	 									
LSUN &77.12  & 62.91  &  76.87   & \textbf{89.21}
    & 94.13 & 86.06  & 96.37  & \textbf{97.89}
	& 89.14 & 80.04 & 92.43 &\textbf{93.48}  \\
									
ImageNet & 79.79 & 62.76   &  85.44  & \textbf{92.97}
    & 94.78 & 85.41 & 97.29 & \textbf{98.37}
	& 90.21 & 79.94 &  93.39 &	\textbf{94.45}\\
									
SCIFAR100 & 76.94  & 48.17   & 86.06   &   \textbf{93.89}
    & 94.18 & 78.94  & 97.43 & \textbf{98.08}
	& 89.57 & 73.72 & 93.02 &	\textbf{95.22}\\

\hline

\end{tabular}
\end{adjustbox}
\label{main_table}
\end{table*}

\indent \textbf{Additional experimental results in the appendix.}  
We also compare area under precision recall curve with SOTA results and report it in the appendix. Logistic regression for assigning weights to the uncertainty scores is trained on a small subset of the iD and OOD samples. We show that these weights can also be learned by only using iD and adversarial samples generated from the iD as a proxy for OODs. All these results, along with ablation studies on indicators of high AU and high EU composing individual detectors as well as individual detectors are included in the appendix. In all the results, we achieved the performance that is similar to the one reported in Table~\ref{main_table}.

\noop{
We demonstrate the effectiveness of the proposed integrated OOD detection method on various datasets such as MNIST~\citep{lenet}, CIFAR10~\citep{CIFAR10}, SVHN~\citep{svhn} and different architectures of the DNN based classifiers for these datasets such as Lenet~\citep{lenet}, ResNet~\citep{resnet} and DenseNet~\citep{densenet}. We measure the following metrics: the true negative rate (TNR) at 95\% true positive rate (TPR), the area under the receiver operating characteristic curve (AUROC), and the detection accuracy (DTACC). 

We compare our results with the baseline softmax prediction probability (SBP)~\citep{baseline}, the ODIN~\citep{odin} and the Mahalanobis method~\citep{mahalanobis} to detect OODs. SBP method exploits the difference in the values of the maximum softmax scores for the ID samples(greater) and OODs(lower) to detect OODs. ODIN tries to further separate out the softmax distributions for the ID and OOD samples by adding small perturbations to the input and applying temperature scaling to the softmax scores. Mahalanobis method exploits the difference in the values of mahalanobis distance of the ID(lower) and OOD(higher) samples from the class-wise mean and tied covariance of the ID samples to detect OODs. 

Due to the space limitation, comparison with the SBP method is listed in the appendix. We also performed experiments to compare our results with the other aforementioned methods on the area under the precision-recall curve for both ID (AUIN) and OOD datasets (AUOUT). Results with this metric are also included in the appendix.

\RK{General highlight applicable to all the in-distribution datasets - Performance gain in the TNR over all the other metrics in all the tested cases due to detection of diverse (instead of uniform) types of OODs.}

\indent \textbf{Experiments with the MNIST dataset.} We trained MNIST on LeNet5. With MNIST as the ID dataset, KMNIST~\citep{kmnist} and Fashion-MNIST(F-MNIST)~\citep{fashion-mnist} were used as the OOD datasets. Table~\ref{table:penul-layer-results} shows that our approach
attains a considerably higher TNR on both the OOD datasets in comparison to the ODIN and the Mahalanobis methods.

\indent \textbf{Experiments with the CIFAR10 dataset.} We trained CIFAR10 on three DNN based classifiers: DenseNet, ResNet34, and ResNet50. With CIFAR10 as the ID dataset, STL10~\citep{stl10}, SVHN~\citep{svhn}, Imagenet~\citep{imagenet}, LSUN~\citep{lsun} and a subset of CIFAR100(SCIFAR100)~\citep{CIFAR10} were used as the OOD datasets. While SVHN, Imagenet and LSUN are different from the CIFAR10 dataset, STL10 and the subset of CIFAR100 OOD datasets used in the experiments are quite close to the ID. We used sea, road, bee and butterfly classes from the CIFAR100 dataset as OODs similar to the ship, automobile and birds classes in the CIFAR10 dataset respectively. 

As shown in Table~\ref{table:penul-layer-results}, the integrated approach performed significantly well on the OODs from SVHN, Imagenet and LSUN datasets on the three classifiers. For instance, it could detect 56\% more OODs as compared to the Mahalanobis method from the SVHN OOD dataset with the Resnet50 classifier and 36\% more OODs as compared to the ODIN method on OODs from the LSUN dataset with Resnet34 classifier. With STL10 and SCIFAR100 datasets, the OODs are expected to be close to the class distributions of the CIFAR10 (ID) dataset resulting in OODs due to epistemic and aleatoric uncertainty in the class-conditional distribution of ID. Although the performance of integrated approach was not significant on STL10 dataset (beacuse of the high TPR of 95\%) but it could detect relatively higher percentage of OODs than the other two methods. With SCIFAR100 dataset, our approach could detect considerably(relatively) higher percentage of OODs than the Mahalanobis(ODIN) method. For instance, it could detect 27\% more OODs than the Mahalanobis method on ResNet50 architecture.

\indent \textbf{Experiments with the SVHN dataset.} We trained SVHN on two DNN based classifiers: DenseNet, and ResNet34. With SVHN as the ID dataset, STL10, CIFAR10, Imagenet, LSUN and, SCIFAR100 were used as the OOD datasets. As shown in Table~\ref{table:penul-layer-results}, our OOD detector could out-perform both the ODIN and the Mahalanobis methods on all the OOD datasets with both the architectures. With 99.61\% TNR on SCIFAR100 and ResNet34 architecture, it could detect 63\% and 13\% more OODs as compared to the ODIN and the Mahalanobis method.
}

\noop{
\RK{Highlight- To perform fair comparison, we performed OOD on feature ensemble(features from multiple layers) with our and Mahalanobis approach. Observations-\\
1) Our approach could boost up OOD detection in all cases, and where there was scope to do so...\\
2) Information from multiple layers is also not sufficient to detect those OODs that a method is unable to capture diverse set of OODs. The lower performance of the Mahalanobis method with the feature ensemble on CIFAR10 with Subset-CIFAR100 (Table~\ref{table:all-layers-results}) as the OOD dataset than the integrated method with just the penultimate layer (Table~\ref{table:penul-layer-results}) gives the evidence for this hypothesis.}
}

\noop{
\indent \textbf{Experiments with feature ensemble.} The results listed in Table~\ref{table:penul-layer-results} are with the Mahalanobis and our method on features from the penultimate layer of the classifier. We also compared performance of our method with the Mahalanobis method on features from all the blocks(layers) of ResNet and DenseNet(LeNet) on CIFAR10 and SVHN(MNIST). This approach is referred as feature ensemble in the Mahalanobis paper~\citep{mahalanobis}. The results of these experiments are listed in Table~\ref{table:all-layers-results}. Here also, our method out-performed the Mahalanobis method on almost all the tested cases. 

An important observation made from these experiments is that information from multiple layers is also not sufficient to detect those OODs that a method is unable to detect. The lower performance of the Mahalanobis method with the feature ensemble on CIFAR10 with SCIFAR100 (Table~\ref{table:all-layers-results}) as the OOD dataset than the integrated method with just the penultimate layer (Table~\ref{table:penul-layer-results}) gives the evidence for this hypothesis. 
}

%% file: conclusion.tex
\section{Conclusion}
\label{sec:conclusion}
We classify the existing techniques as detecting OODs due to either high EU or high AU. We demonstrate that these techniques differ in their ability for OOD detection. Using these insights, we propose using an ensemble approach for detecting OODs as datapoints with high uncertainty (aleatoric or epistemic).
We have performed extensive experiments on a toy dataset and several benchmark datasets (e.g., MNIST, CIFAR10, SVHN).
Our experiments show that our approach can accurately detect various types of OODs coming from a wide range of OOD datasets. We have shown that our approach generalizes over multiple DNN architectures and performs robustly when the OOD samples are similar to iD. 

The difference in the ability of individual detectors could be explained with their expertise in detecting particular types of OODs. Mixture of experts model (MoE)~\citep{moe} is used to make each expert focus on predicting the right answer for the cases where it is already doing better than the other experts. As a future work, we will look into MoE for dynamically (i.e. conditioned on input) assigning weights to individual detectors in the ensemble approach.


%% file: ack.tex
\subsubsection*{Acknowledgments}
This work was supported by the Air Force Research Laboratory and the Defense Advanced Research Projects Agency under 
DARPA Assured Autonomy under 
Contract No. FA8750-19-C-0089 and Contract No. FA8750-18-C-0090, 
U.S. Army Research Laboratory Cooperative Research Agreement  W911NF-17-2-0196, 
U.S. National Science
Foundation(NSF) grants \#1740079 and \#1750009,
the Army Research Office under Grant Number W911NF-20-1-0080 and in part by Semiconductor Research Corporation (SRC) Automotive Electronics program under Task 2894.001. Any opinions, findings and conclusions or recommendations expressed in this material are those of the authors and do not necessarily reflect the views of the Air Force Research Laboratory (AFRL), the Army Research Office (ARO), the Defense Advanced Research Projects Agency (DARPA), or the Department of Defense, or the United States Government.

%% file: appendix.tex
\appendix
\section{Appendix}
\label{appendix}
\subsection{Existing OOD detection Techniques}
Existing techniques for detecting OODs due to either high AU or high EU are summarized in Table~\ref{tab:detect_oods_eu_au}.
\begin{table*}[]
    \centering
    \begin{adjustbox}{width=2\columnwidth,center}
    \begin{tabular}{|c|c|}
        \hline
        Detection of OODs due to high AU & Justification \\
        \hline
        ODIN~\citep{odin} & ODIN is an enhancement to SBP after adding noise to the input  \\ & and temperature scaling to the classifier's confidence. \\
        DkNN~\citep{dknn} & Use entropy in the labels of the kNNs to detect OODs.\\
        Confident Classifier~\citep{kl_div_entropy} & Train the OOD detector by minimizing KL-divergence between  the predicted \\ & distribution of the softmax scores and uniform distribution for OODs.\\
        Self-supervised~\citep{self-supervised} & Use KL-divergence from the uniform distribution \\ & of the predicted softmax scores to detect OODs.\\
        Outlier Exposure~\citep{oe} & Train the OOD detector by setting cross-entropy loss \\ & for OODs ($\mathcal L_{OE}$) as the uniform distribution.\\
        Predictive Uncertainty~\citep{proper_scoring} & Use entropy in the predicted class distributions for OOD detection.\\ 
        \hline
        Detection of OODs due to high EU  & Justification \\
        \hline
        OC-SVM~\citep{oc-svm} & Lack of support from the estimated iD density is used for OOD detection.\\
        Deep-SVDD~\citep{deep-svdd} &  Lack of support from the estimated iD \\ & density (as a hypersphere) is used for OOD detection.\\
        VAE~\citep{vae-recon-err} & Reconstruction error is used to detect OODs. The high error \\ & for OODs is due to lack of support by the iD data as the VAE is \\ & trained to reconstruct only the iD data.\\
        Outlier Exposure~\citep{oe} &  Train the OOD detector by setting the loss \\ & function based on density estimation from the iD.\\
        GEOM~\citep{geometric} & Error in the prediction of the applied transformation  on an input \\ GOAD~\citep{GOAD} &  is used to detect OODs. The high error in the prediction for OODs  is due to \\ & lack of support from iD as this task is learnt by transforming only the iD data. \\
        \hline
    \end{tabular}
    \end{adjustbox}
    \caption{Existing techniques detecting OODs due to high EU or high AU.}
    \label{tab:detect_oods_eu_au}
\end{table*}



\subsection{Novel non-conformance measure amongst the nearest neighbors for detecting OODs.}
We compute an m-dimensional feature vector to capture the conformance among the input's nearest neighbors in the training samples, where m is the dimension of the input. We call this m-dimensional feature vector as the conformance vector. The conformance vector is calculated by taking the mean deviation along each dimension of the nearest neighbors from the input. We hypothesize that this deviation for the iD samples would vary from the OODs due to AU; i.e. uncertainty due to nearest neighbors from multiple classes in case of OODs.

The value of the conformance measure is calculated by computing mahalanobis distance of the input's conformance vector to the closest class conformance distribution. The parameters of this mahalanobis distance are the empirical class means and tied empirical covariance on the conformance vectors of the training samples. 

The value of the number of the nearest neighbors is chosen from the set $\{10, 20, 30, 40, 50\}$ via validation. 
We used Annoy (Approximate Nearest Neighbors Oh Yeah)~\citep{annoy} to compute the nearest neighbors.

\subsection{Additional experimental results}
We first present comparison of AUPR with SOTA.
We then present our results on various vision datasets and different architectures of the pre-trained  DNN based classifiers for these datasets in comparison to the SBP, ODIN, the Mahalanobis methods in unsupervised settings. Finally, we report results from the ablation study on OOD detection with indicators of high EU (Mahala, PCA) and AU (ODIN, KNN) as well as individual detectors (Mahala+ODIN, PCA+KNN) and compare it the proposed ensemble approach.

\subsubsection{Comparison of AUPR with the SOTA.}
\label{apdx:aupr}
Results in comparison to AUPR IN and AUPR OUT are shown in tables~\ref{table:mnist-aupr},
~\ref{table:cifar10-densenet},~\ref{table:cifar10-resent34},~\ref{table:cifar10-resent50},~\ref{table:svhn-densenet}, and~\ref{table:svhn-resent34}. Here also, the proposed OOD detection technique could out-perform the other three detectors on most of the test cases. 

\subsubsection{Learning weights of the individual detectors in unsupervised settings}
\label{apdx:unsuper}
Here we train the logistic regression on a small subset of iD and a small subset of adversarial samples as a proxy for OODs for determining weights of the two detectors. Adversarial samples are generated by applying FGSM attack~\citep{fgsm} on the iD samples. Table~\ref{table:results} shows that the proposed approach works in the unsupervised settings as well.




\begin{table*}
\caption{Experimental Results with MNIST on Lenet5 for AUPR IN and AUPR OUT. \\ The best results are highlighted.}
\begin{center}
\begin{tabular}{llllllll}
\hline
\multicolumn{1}{c}{\bf OOD Dataset}   &\multicolumn{1}{c}{\bf Method} &\multicolumn{1}{c}{\bf AUPR IN} &\multicolumn{1}{c}{\bf AUPR OUT}

\\ \hline \\
KMNIST       &SBP          	&92.47	&92.41 \\
             &ODIN             	&92.65	&92.69\\
             &Mahalanobis       	&96.69	&96.2 \\
             &Ours                 &\textbf{98.47}	&\textbf{98.12}\\ 
\\ \hline \\
Fashion-MNIST      &SBP       &87.98	 &87.89\\
                   &ODIN           &90.94	 &89.99\\
                   &Mahalanobis  	&95.24	 &91.94\\
                   &Ours           &\textbf{96.53}	&\textbf{93.04}\\

\\ \hline 

\end{tabular}
\end{center}
\label{table:mnist-aupr}
\end{table*}

\begin{table*}
\caption{Experimental Results with CIFAR10 on DenseNet for AUPR IN and AUPR OUT. The best results are highlighted.}
\begin{center}
\begin{tabular}{llllllll}
\hline
\multicolumn{1}{c}{\bf OOD Dataset}  
&\multicolumn{1}{c}{\bf Method} &\multicolumn{1}{c}{\bf AUPR IN} &\multicolumn{1}{c}{\bf AUPR OUT}
\\ \hline \\
SVHN      &SBP        &74.53	&94.09\\
          &ODIN            &80.49	&97.05\\
          &Mahalanobis 	&94.13	&98.78\\
          &Ours           &\textbf{96.27}	&\textbf{99.39}\\
       \\ \hline \\
Imagenet     &SBP &90.88	&86.74\\
             &ODIN  	&91.32	&90.55\\
             &Mahalanobis &91.32	&88.6\\
             &Ours 	&\textbf{96.08}	&\textbf{95.56}\\
          \\ \hline \\
LSUN      &SBP 	&93.68	&89.83\\
         &ODIN     	&94.65 &93.39\\
         &Mahalanobis 	&92.71 &87.74\\
          &Ours 	&\textbf{97.40}	&\textbf{96.51}\\
       \\ \hline \\
Subset CIFAR100    &SBP 	&96.65	&50.08 \\
                   &ODIN 	&95.14	&47.64\\
                   &Mahalanobis 	&95.68	&37.86 \\
                   &Ours &\textbf{98.15}	&\textbf{52.18}\\
\\ \hline 
\end{tabular}
\end{center}
\label{table:cifar10-densenet}
\end{table*}

\begin{table*}
\caption{Experimental Results with CIFAR10 on ResNet34 for AUPR IN and AUPR OUT. The best results are highlighted.}
\begin{center}
\begin{tabular}{llllllll}
\hline
\multicolumn{1}{c}{\bf OOD Dataset}   &\multicolumn{1}{c}{\bf Method} &\multicolumn{1}{c}{\bf AUPR IN} &\multicolumn{1}{c}{\bf AUPR OUT}
\\ \hline \\
   
SVHN      &SBP 	&85.4	&93.96\\
          &ODIN  	&86.46	&97.55\\
          &Mahalanobis	&91.19	&96.14\\
          & Ours  	&\textbf{93.61}	&\textbf{98.67}\\
 \\ \hline \\ 
Imagenet    &SBP  	&92.49	&88.4 \\
         &ODIN    	&92.11	&87.46\\
          &Mahalanobis    	&95.77	&94.02\\
          & Ours   	&\textbf{96.32}	&\textbf{94.99}\\
        \\ \hline \\ 
LSUN      &SBP	&92.45	&88.55\\
          &ODIN     	&91.58	&86.5\\
          &Mahalanobis 	&97.08	&95.78\\
          & Ours  &\textbf{97.36}	&\textbf{96.29}\\
        \\ \hline \\ 
Subset CIFAR100    &SBP   	&97.77	&55.62\\
                   &ODIN  	&97.05	&51.57\\
                   &Mahalanobis 	&97.71	&54.11\\
                   & Ours 	&\textbf{98.91}	&\textbf{60.11}\\
\\ \hline \\
\end{tabular}
\end{center}
\label{table:cifar10-resent34}
\end{table*}

\begin{table*}
\caption{Experimental Results with CIFAR10 on ResNet50 for AUPR IN and AUPR OUT. The best results are highlighted.}
\begin{center}
\begin{tabular}{llllllll}
\hline
\multicolumn{1}{c}{\bf OOD Dataset}   &\multicolumn{1}{c}{\bf Method} &\multicolumn{1}{c}{\bf AUPR IN} &\multicolumn{1}{c}{\bf AUPR OUT}

                \\ \hline \\
SVHN               &SBP    	&87.78	&95.61 \\
                   &ODIN        	&93.17	&99.03 \\
                   &Mahalanobis 	&71.88	&92.54   \\
                   & Ours       	&\textbf{94.69}	&\textbf{99.20}\\
                \\ \hline \\
Imagenet           &SBP    	&92.6	&87.98 \\
                   &ODIN        	&95.16	&94.45\\
                   &Mahalanobis 	&86.14	&80.6   \\
                   & Ours      	&\textbf{96.11}	&\textbf{94.85} \\
                                \\ \hline \\
LSUN               &SBP     &94.45	 &90.41 \\
                   &ODIN         	&96.9	 &96.01\\
                   &Mahalanobis   	&89.34	 &82.87  \\
                   & Ours         	 &\textbf{97.53}	 &\textbf{96.03}\\
                                \\ \hline \\
Subset CIFAR100   &SBP     	&97.72	&55.29 \\
                   &ODIN         	&96.67	&\textbf{60.62} \\
                   &Mahalanobis  	&94.49	&36.12  \\
                   & Ours        	&\textbf{98.72}	&59.30\\
\\ \hline

\end{tabular}
\end{center}
\label{table:cifar10-resent50}
\end{table*}

\begin{table*}
\caption{Experimental Results with SVHN on DenseNet for AUPR IN and AUPR OUT. The best results are highlighted.}
\begin{center}
\begin{tabular}{llllllll}
\hline
\multicolumn{1}{c}{\bf OOD Dataset}   &\multicolumn{1}{c}{\bf Method} &\multicolumn{1}{c}{\bf AUPR IN} &\multicolumn{1}{c}{\bf AUPR OUT}

\\ \hline \\

CIFAR10     &SBP   &95.7	 &82.8   \\
        &ODIN      &84.32 &	60.32\\
          &Mahalanobis  	 &98.94	 &88.91\\
          & Ours	&\textbf{99.04}	&\textbf{90.54}\\
       \\ \hline \\
Imagenet    &SBP    &97.2	 &88.42\\
          &ODIN     &90.95	 &79.59\\
          &Mahalanobis  	 &99.12	 &90.22\\
          & Ours  	&\textbf{99.4}	&\textbf{95.17}\\
       \\ \hline \\
LSUN      &SBP      &96.96	 &87.44\\
          &ODIN      &92.03	 &79.98\\
          &Mahalanobis      &98.84	 &85.79\\
          & Ours   &\textbf{99.26}	&\textbf{93.34}\\
       \\ \hline \\
Subset CIFAR100    &SBP   	 &99.39	 &63.21\\
                   &ODIN     &97.24	 &45.23\\
                   &Mahalanobis     	 &99.82	 &\textbf{72.35}\\
                  & Ours   	&\textbf{99.86}	&70.10\\

\\ \hline 
\end{tabular}
\end{center}
\label{table:svhn-densenet}
\end{table*}

\begin{table*}
\caption{Experimental Results with SVHN on ResNet34 for AUPR IN and AUPR OUT. The best results are highlighted.}
\begin{center}
\begin{tabular}{llllllll}
\hline
\multicolumn{1}{c}{\bf OOD Dataset}   &\multicolumn{1}{c}{\bf Method} &\multicolumn{1}{c}{\bf AUPR IN} &\multicolumn{1}{c}{\bf AUPR OUT}

\\ \hline \\

CIFAR10     &SBP 	&95.06	&85.66\\
        &ODIN        	&80.69	&50.49\\
          &Mahalanobis 	&99.04	&88.62\\
          & Ours       &\textbf{99.17}	&\textbf{91.17}\\
       \\ \hline \\
Imagenet    &SBP  &95.68	&86.18\\
          &ODIN        &84.62	&58.28\\
          &Mahalanobis &99	&88.39\\
          & Ours      &\textbf{99.19}	&\textbf{90.78}\\
       \\ \hline \\
LSUN      &SBP    &94.19	&83.95\\
          &ODIN        &82.37	&53.12\\
          &Mahalanobis &98.73	&85.11\\
          & Ours       &\textbf{99.03}	&\textbf{89.03}\\
       \\ \hline \\
Subset CIFAR100        &SBP    &99.35	&64.38\\
          &ODIN        &95.57	&23.04\\
          &Mahalanobis &99.81	&64.4\\
          & Ours      &\textbf{99.88}	&\textbf{65.19}

\\ \hline 
\end{tabular}
\end{center}
\label{table:svhn-resent34}
\end{table*}

\subsubsection{Ablation study}
\label{apdx:ablation}
\textbf{Indicators of high EU or high AU.} We report ablation study on OOD detection with the following indicators of high EU and high AU composing the individual detectors. Mahala is used as an indicator of high EU in the first detector, ODIN used as an indicator of high AU in the first detector, PCA used as an indicator of high EU in the second detector, and KNN used as an indicator of high AU in the second detector. Tables~\ref{table:ablation-cifar},~\ref{table:ablation-svhn-densenet} and~\ref{table:cifar10-densenet2} show these results. 

The proposed approach could out-perform all the four OOD detection techniques in all the cases. An important observation made from these experiments is that the performance of OOD detection techniques based on high EU or high AU could depend on the architecture of the classifier.  For example, while the performance of PCA was really bad in case of DenseNet (for both CIFAR10 and SVHN) as compared to all other methods, it could out-perform all but our approach for SVHN on ResNet34.

\textbf{Individual detectors.} We also report the performance of individual detectors (Mahala+ODIN and PCA+KNN) used in the ensemble approach. The uncertainty score by these detectors is used for OOD detection. Table~\ref{table:resnet34_ind_detectors} shows that the ensemble approach could out-perform individual detectors in almost all the cases.

\begin{table*}
\begin{center}
\caption{Comparison with SBP, ODIN and Mahalanobis methods in unsupervised settings.}
\begin{tabular}{ccccccc}
\hline \\


\multicolumn{1}{c}{\bf In-dist} & \multicolumn{1}{c}{\bf OOD Dataset}  &\multicolumn{1}{c}{\bf Method} &\multicolumn{1}{c}{\bf TNR} &\multicolumn{1}{c}{\bf AUROC} & \multicolumn{1}{c}{\bf DTACC} & \multicolumn{1}{c}{\bf AUPR}
\\ \multicolumn{1}{c}{\bf (model)} 


\\ \hline

          \\ CIFAR10\\
          (ResNet50)\\

\\ \hline \\

      & SVHN      & SBP &  44.69	 &\textbf{97.31} 	& 86.36 & 87.78  \\
      &           & ODIN & 63.57 & 93.53 & 86.36 & 87.58 \\
      &           & Mahalanobis & 72.89 & 91.53 & 85.39 & 73.80   \\
      &           & Ours & \textbf{85.90}  &95.14  &\textbf{90.66}  &\textbf{80.01}   \\
      & Imagenet  & SBP &  42.06 	 &90.8	& 84.36 & 92.6  \\
      &           & ODIN & 79.48 & 96.25 &  90.07 & \textbf{96.4}5    \\
      &           & Mahalanobis  &94.26  & \textbf{97.41} & 95.16 & 93.11   \\
      &           & Ours   &      \textbf{95.19} & 97.00 & \textbf{96.02} &90.92    \\
               
      & LSUN  & SBP &  48.37 	 &92.78	& 86.97 & 94.45   \\
      &           & ODIN  & 87.29  & 97.77  & 92.65 &  97.96  \\
      &           & Mahalanobis  & 98.17 & 99.38 & 97.38 & 98.69   \\
      &           & Ours &\textbf{99.36} & \textbf{99.65} & \textbf{98.57} & \textbf{98.96}  \\
\\ \hline

       \\CIFAR10\\
       (WideResNet)\\  
\\ \hline \\

      & SVHN      & SBP  & 45.46 &  90.10 & 82.91 &  82.52  \\ 
      &           & ODIN & 57.14  & 89.30  & 81.14 &  75.48  \\
      &           & Mahalanobis & 85.86 & 97.21 &  91.87 &  \textbf{94.69}  \\
      &           & Ours & \textbf{88.95}  & \textbf{97.61}  & \textbf{92.46} & 92.84  \\
      & LSUN      & SBP & 52.64 & 92.89 & 86.81  & 94.13\\
      &           & ODIN & 79.60 & 96.08 &  89.74  & 96.23 \\
      &           & Mahalanobis & 95.69 & 98.93  & 95.41 & 98.99   \\
      &           & Ours & \textbf{98.84}  & \textbf{99.63} & \textbf{97.72} & \textbf{99.25}\\

\\ \hline\\
         SVHN\\
         (DenseNet)\\
\\ \hline \\
      & Imagenet  & SBP &  79.79 	 &94.78	& 90.21 &97.2 \\
      &           & ODIN & 79.8 &  94.8 & 90.2 & 97.2  \\
      &           & Mahalanobis & \textbf{99.85} &  \textbf{99.88} & \textbf{98.87} & \textbf{99.95}\\
      &           & Ours &   98.02      	 & 98.34	&  98.00 &97.05 \\
      & LSUN  & SBP &  77.12 	 &94.13	& 89.14 &96.96   \\
      &           & ODIN & 77.1 & 94.1 & 89.1 & 97.0\\
      &           & Mahalanobis  & \textbf{99.99} & \textbf{99.91} & \textbf{99.23} & \textbf{99.97}\\
      &           & Ours    	 & 99.74  & 99.79  & 99.08 &99.65   \\
            & CIFAR10   & SBP &  69.31 	 &91.9 	& 86.61  & 95.7 \\
      &           & ODIN & 69.3 & 91.9 & 86.6 & 95.7  \\
      &           & Mahalanobis &  \textbf{97.03} & \textbf{98.92} & \textbf{96.11} & \textbf{99.61}\\
      &           & Ours & 94.87  & 98.41  & 94.97& 98.76  \\
      
\\ \hline \\      
\end{tabular}
\label{table:results}
\end{center}
\end{table*}

\begin{table*}

\caption{Ablation study with individual indicators of uncertainty (either AU or EU) for CIFAR10 on DenseNet. \\ The best results are highlighted.}
\begin{center}
\begin{tabular}{ccccccc}
\label{table:ablation-cifar}
\\ \hline \\
\multirow{2}{*}{}  {\bf OOD} & {\bf Method} & {\bf TNR} & {\bf AUROC} & {\bf DTACC} & {\bf AUPR} & {\bf AUPR} \\
 {\bf dataset} &  & {\bf (TPR=95\%)} &  & &{\bf IN} &{\bf OUT}\\
\\ \hline \\

SVHN   &   Mahala       &83.63	&97.1	&91.26	&94.13	&98.78\\
       &   KNN 	      &84.07	&97.18	&91.32	&94.2	&98.84\\
       &   ODIN	              &69.96	&92.02	&84.1	&80.49	&97.05\\
       &   PCA                &2.46	    &55.89	&56.36	&35.42	&74.12 \\
       &   Our         &\textbf{90.92}	&\textbf{98.41}	&\textbf{93.29}	&\textbf{96.27}	&\textbf{99.39}\\
       \\ \hline \\
Imagenet & Mahala &49.33	&90.32	&83.08	&91.32	&88.6\\
       &   KNN     &51.36	&90.73	&83.31	&91.75	&88.87\\
       &   ODIN	        &61.03	&91.4	&83.85	&91.32	&90.55\\
       &   PCA          &4.66	&58.68	&57.19	&60.66	&54.42 \\
       &   Ours           &\textbf{77.56}	&\textbf{95.86}	&\textbf{89.55}	&\textbf{96.08}	&\textbf{95.56}\\
       				
          \\ \hline \\
LSUN   &   Mahala  &46.63	&91.18	&84.93	&92.71	&87.74\\
       &   KNN 	&51.48	&92.25	&85.96	&93.75	&89.13\\
       &   ODIN	        &71.89	&94.37	&87.72	&94.65	&93.39\\
       &   PCA          &2.06	&53.26	&54.88	&57.08	&49.33\\
       &   Ours           &\textbf{83.47}	&\textbf{97.07}	&\textbf{91.74}	&\textbf{97.4}	&\textbf{96.51}\\
       \\ \hline \\
\end{tabular}
\end{center}
\label{table:cifar10-densenet1}
\end{table*}

\begin{table*}
\caption{Ablation study with individual indicators of uncertainty (either AU or EU) for SVHN on DenseNet. \\ The best results are highlighted.}
\begin{center}
\begin{tabular}{ccccccc}
\hline \\
\multirow{2}{*}{}  {\bf OOD} & {\bf Method} & {\bf TNR} & {\bf AUROC} & {\bf DTACC} & {\bf AUPR} & {\bf AUPR} \\
 {\bf dataset} &  & {\bf (TPR=95\%) }&  & &{\bf IN} &{\bf OUT}\\
\\ \hline \\

CIFAR10   &   Mahala     &80.82	&96.8	&92.27	&98.94	&88.91\\
       &   KNN 	      &69.99	&95.58	&90.77	&98.52	&84.3\\
       &   ODIN	              &37.23	&73.14	&68.92	&84.32	&60.32\\
       &   PCA                &5.27	&65.82	&64.83	&86.62	&33.51 \\
       &   Ours         &\textbf{83.81}	&\textbf{97.17}	&\textbf{92.69}	&\textbf{99.04}	&\textbf{90.54}\\
       \\ \hline \\
Imagenet   &   Mahala &85.44	&97.29	&93.39	&99.12	&90.22\\
       &   KNN 	  &65.76	&94.67	&89.59	&98.18	&80.16\\
       &   ODIN	          &62.76	&85.41	&79.94	&90.95	&79.59\\
       &   PCA            &5.16 	&65.08	&65.39	&86.65	&32.83\\
       &   Ours           	&\textbf{92.97}	&\textbf{98.37}	&\textbf{94.45}	&\textbf{99.40}	&\textbf{95.17}\\
       \\ \hline \\
LSUN   &   Mahala    &76.87	&96.37	&92.43	&98.84	&85.79\\
       &   KNN     &59.64	&93.71	&88.22	&97.83	&77.17\\
       &   ODIN	        &62.91	&86.06	&80.04	&92.03	&79.98\\
       &   PCA          &3.19	&62.66	&64.7	&85.72	&30.37 \\
       &   Ours           &\textbf{89.21}	 &\textbf{97.89}	 &\textbf{93.48}	 &\textbf{99.26}	 &\textbf{93.34}
\\
          \\ \hline \\
\end{tabular}
\end{center}
\label{table:ablation-svhn-densenet}
\end{table*}
		 
\begin{table*}
\caption{Ablation study with individual indicators of uncertainty (either AU or EU) for SVHN on ResNet34. \\ The best results are highlighted.}
\begin{center}
\begin{tabular}{ccccccc}
\hline \\
\multirow{2}{*}{}  {\bf OOD} & {\bf Method} & {\bf TNR} & {\bf AUROC} & {\bf DTACC} & {\bf AUPR} & {\bf AUPR} \\
 {\bf dataset} &  & {\bf (TPR=95\%)} &  & &{\bf IN} &{\bf OUT}\\
\\ \hline \\

SCIFAR100   &   Mahala   &86.61	&97.3	&93.6	&99.81	&64.4\\
       &   KNN 	      &84.67	&96.82	&92.83	&99.76	&61.08\\
       &   ODIN	              &36.67	&68.01	&67.26	&95.57	&23.04\\
       &   PCA                &89.94	&97.81	&94.52	&99.84	&70.83 \\
       &   Ours         &\textbf{97.28}	&\textbf{98.19}	&\textbf{96.39}	&\textbf{99.88}	&\textbf{65.19}\\
       \\ \hline \\
LSUN &   Mahala    &78.38	&96.17	&91.98	&98.73	&85.11\\
       &   KNN     &77.61	&95.98	&91.34	&98.61	&85.56\\
       &   ODIN	        &35.92	&68.6	&66.75	&82.37	&53.12\\
       &   PCA          &82.93	&96.88	&92.74	&98.97	&88.27 \\
       &   Ours           &\textbf{85.46}	&\textbf{97.09}	&\textbf{93.17}	&\textbf{99.03}	&\textbf{89.03}\\
          \\ \hline \\
CIFAR10   &   Mahala &85.03	&97.05	&93.15	&99.04	&88.62\\
       &   KNN 	  &82.17	&96.65	&92.24	&98.87	&87.63\\
       &   ODIN	          &32.67	&66.75	&65.37	&80.69	&50.49\\
       &   PCA            &88.18	&97.55	&93.83	&99.2	&90.77\\
       &   Ours           &\textbf{90.34}	&\textbf{97.64}	&\textbf{94.29}	&\textbf{99.17}	&\textbf{91.17}\\
       \\ \hline \\
\end{tabular}
\label{table:ablation-svhn}
\end{center}
\label{table:cifar10-densenet2}
\end{table*}

\begin{table*}
\caption{Ablation study on individual detectors with uncertainty scores on ResNet34. \\ The best results are highlighted.}
\begin{center}
\begin{tabular}{cccccccc}
\hline \\
\multirow{2}{*}{} {\bf in-dist} & {\bf OOD} & {\bf Method} & {\bf TNR} & {\bf AUROC} & {\bf DTACC} & {\bf AUPR} & {\bf AUPR} \\
& {\bf dataset} &  & {\bf (TPR=95\%)} &  & &{\bf IN} &{\bf OUT}\\
\\ \hline \\

CIFAR10 & SVHN &   Mahala+ODIN &  76.98 &	96.09	& 90.24	& 92.73 &	98.17\\
       &  & PCA+KNN	  & 65.82    &94.92	& 90.09	 & 92.26 &	96.79\\
       &  & Ours         &\textbf{83.20}	& \textbf{96.91}	 & \textbf{91.16} &	\textbf{93.61} &	\textbf{98.67}\\
       & LSUN &   Mahala+ODIN   &79.58 &	96.59 &	90.88 &	97.07	& 95.99\\
       &  & PCA+KNN	      &75.54 &	96.30	& 90.58 &	96.96 &	95.45\\
       &  & Ours         &\textbf{81.23} &	\textbf{96.87} &\textbf{91.19}	&\textbf{97.36}	&\textbf{96.29}\\
       & Imagenet &   Mahala+ODIN  & 73.18 &	95.61 &	89.57 &	96.17 &	94.86\\
       &  & PCA+KNN	      &67.47 &	94.76 &	88.50 &	95.55 &	93.63\\
       &  & Ours         &\textbf{74.53} &	\textbf{95.73} &	\textbf{89.73}	&\textbf{96.32}	&\textbf{94.99}\\
       & SCIFAR100 &   Mahala+ODIN   &47.50 &	91.86 &	85.84	& 98.43 &	60.82\\
       &  & PCA+KNN	      &50.28 &	92.82 &	87.82 &	98.67	& 59.91\\
       &  & Ours         &\textbf{51.11}	&\textbf{93.85} &	\textbf{89.93}	&\textbf{98.91} &	\textbf{60.11}\\
       \\ \hline \\
SVHN & CIFAR10 &   Mahala+ODIN   &86.32 &	97.06 &	93.47 &	99.00 & 88.62\\
       &  & PCA+KNN	      &88.62 &	97.57 &	93.84 &	\textbf{99.20} &	90.87\\
           &  & Ours         &\textbf{90.34} &	\textbf{97.64}&	\textbf{94.29}	&99.17 &\textbf{91.17}\\
       & LSUN &   Mahala+ODIN   &79.00 &	96.19 &	92.14 &	98.73 &	85.01\\
       &  & PCA+KNN	      &84.49 &	96.96 &	93.10 & 98.99 &	88.27\\
       &  & Ours         &\textbf{85.46}  &\textbf{97.09}  &\textbf{93.17}  &	\textbf{99.03}  &\textbf{89.03}\\
       & Imagenet &   Mahala+ODIN   &84.59 &	96.94 &	93.23 &	98.99 &	88.26\\
       &  & PCA+KNN	      &88.58 &	97.55 &	93.95 &	\textbf{99.20} &	\textbf{90.81}\\
       &  & Ours         &\textbf{89.81} &\textbf{97.60} &\textbf{94.32} &99.19 &90.78\\
       & SCIFAR100 &   Mahala+ODIN   &86.56 &	97.32 &	93.75 &	99.81 &	64.67\\
       &  & PCA+KNN	      &95.72 & 98.14 &	96.01 &	99.87 &	\textbf{66.46}\\
       &  & Ours         &\textbf{97.28}	& \textbf{98.19} &	\textbf{96.39} &	\textbf{99.88}	& 65.19\\
       \\ \hline \\
\end{tabular}
\label{table:ablation-svhn}
\end{center}
\label{table:resnet34_ind_detectors}
\end{table*}


\noop{
\newpage
\subsection{Intuition on the failure of different methods}
DKNN fig~\ref{fig:dknn_counter}
\begin{figure}[h]
\centering
\includegraphics[width=1\textwidth]{figures/apprendix/dknn_counter_intuition.png}
\caption{OODs with the conformance measure in the labels of the nearest neighbors similar to the in-distribution samples, as visualized in 2D space}
\label{fig:dknn_counter}
\end{figure}

Mahalanobis fig~\ref{fig:mahalanobis_counter}, plot with the least probable OODs generated from regressor trained with Mahalanobis distance
\begin{figure}[h]
\centering
\includegraphics[width=1\textwidth]{figures/apprendix/mahala_counter_intuition.png}
\caption{OODs with mahalanobis distance of their features from the penultimate layer close to the corresponding mahalanobis distance of the in-distribution samples, as visualized in the 2D space}
\label{fig:mahalanobis_counter}
\end{figure}

Calibration by temp scaling/ODIN/Baseline/Softmax fig~\ref{fig:baseline_counter}, generated with the spiral data and a classifier trained on this data
\begin{figure}[h]
\centering
\includegraphics[width=1\textwidth]{figures/apprendix/baseline_counter_intuition.png}
\caption{OODs lying within the classification boundary of a particular class. Similar to the in-distribution samples, the softmax score of these OODs is very high ($\geq 0.95$).}
\label{fig:baseline_counter}
\end{figure}

LID fig~\ref{fig:lid_counter}
\begin{figure}[h]
\centering
\includegraphics[width=1\textwidth]{figures/apprendix/lid_counter_intuition.png}
\caption{OODs with local intrinsic dimensionality measure close to the in-distribution samples}
\label{fig:lid_counter}
\end{figure}

PCA fig~\ref{fig:pca_counter}
\begin{figure}[h]
\centering
\includegraphics[width=1\textwidth]{figures/apprendix/pca_counter_intuition.png}
\caption{OODs lying along the principal component of the in-distribution samples}
\label{fig:pca_counter}
\end{figure}
}

%% file: main.bbl
\begin{thebibliography}{34}
\providecommand{\natexlab}[1]{#1}
\providecommand{\url}[1]{\texttt{#1}}
\expandafter\ifx\csname urlstyle\endcsname\relax
  \providecommand{\doi}[1]{doi: #1}\else
  \providecommand{\doi}{doi: \begingroup \urlstyle{rm}\Url}\fi

\bibitem[An \& Cho(2015)An and Cho]{vae-recon-err}
An, J. and Cho, S.
\newblock Variational autoencoder based anomaly detection using reconstruction
  probability.
\newblock \emph{Special Lecture on IE}, 2\penalty0 (1):\penalty0 1--18, 2015.

\bibitem[Bergman \& Hoshen(2020)Bergman and Hoshen]{GOAD}
Bergman, L. and Hoshen, Y.
\newblock Classification-based anomaly detection for general data.
\newblock \emph{arXiv preprint arXiv:2005.02359}, 2020.

\bibitem[Bernhardsson(2018)]{annoy}
Bernhardsson, E.
\newblock Annoy, 2018.
\newblock URL \url{https://github.com/spotify/annoy}.

\bibitem[Bojarski et~al.(2016)Bojarski, Del~Testa, Dworakowski, Firner, Flepp,
  Goyal, Jackel, Monfort, Muller, Zhang, et~al.]{ML-App7}
Bojarski, M., Del~Testa, D., Dworakowski, D., Firner, B., Flepp, B., Goyal, P.,
  Jackel, L.~D., Monfort, M., Muller, U., Zhang, J., et~al.
\newblock End to end learning for self-driving cars.
\newblock \emph{arXiv preprint arXiv:1604.07316}, 2016.

\bibitem[Clanuwat et~al.(2018)Clanuwat, Bober-Irizar, Kitamoto, Lamb, Yamamoto,
  and Ha]{kmnist}
Clanuwat, T., Bober-Irizar, M., Kitamoto, A., Lamb, A., Yamamoto, K., and Ha,
  D.
\newblock Deep learning for classical japanese literature.
\newblock \emph{arXiv preprint arXiv:1812.01718}, 2018.

\bibitem[De~Fauw et~al.(2018)De~Fauw, Ledsam, Romera-Paredes, Nikolov, Tomasev,
  Blackwell, Askham, Glorot, O’Donoghue, Visentin, et~al.]{NNclinically}
De~Fauw, J., Ledsam, J.~R., Romera-Paredes, B., Nikolov, S., Tomasev, N.,
  Blackwell, S., Askham, H., Glorot, X., O’Donoghue, B., Visentin, D., et~al.
\newblock Clinically applicable deep learning for diagnosis and referral in
  retinal disease.
\newblock \emph{Nature medicine}, 24\penalty0 (9):\penalty0 1342--1350, 2018.

\bibitem[Deng et~al.(2009)Deng, Dong, Socher, Li, Li, and Fei-Fei]{imagenet}
Deng, J., Dong, W., Socher, R., Li, L.-J., Li, K., and Fei-Fei, L.
\newblock Imagenet: A large-scale hierarchical image database.
\newblock In \emph{2009 IEEE conference on computer vision and pattern
  recognition}, pp.\  248--255. Ieee, 2009.

\bibitem[Dietterich(2000)]{ensemble}
Dietterich, T.~G.
\newblock Ensemble methods in machine learning.
\newblock In \emph{International workshop on multiple classifier systems}, pp.\
   1--15. Springer, 2000.

\bibitem[Gkioxari et~al.(2015)Gkioxari, Girshick, and
  Malik]{img-classification}
Gkioxari, G., Girshick, R., and Malik, J.
\newblock Contextual action recognition with r* cnn.
\newblock In \emph{Proceedings of the IEEE international conference on computer
  vision}, pp.\  1080--1088, 2015.

\bibitem[Golan \& El-Yaniv(2018)Golan and El-Yaniv]{geometric}
Golan, I. and El-Yaniv, R.
\newblock Deep anomaly detection using geometric transformations.
\newblock In \emph{Advances in Neural Information Processing Systems}, pp.\
  9758--9769, 2018.

\bibitem[Goodfellow et~al.(2014)Goodfellow, Shlens, and Szegedy]{fgsm}
Goodfellow, I.~J., Shlens, J., and Szegedy, C.
\newblock Explaining and harnessing adversarial examples.
\newblock \emph{arXiv preprint arXiv:1412.6572}, 2014.

\bibitem[Guo et~al.(2017)Guo, Pleiss, Sun, and Weinberger]{guo2017calibration}
Guo, C., Pleiss, G., Sun, Y., and Weinberger, K.~Q.
\newblock On calibration of modern neural networks.
\newblock \emph{arXiv preprint arXiv:1706.04599}, 2017.

\bibitem[Hannun et~al.(2014)Hannun, Case, Casper, Catanzaro, Diamos, Elsen,
  Prenger, Satheesh, Sengupta, Coates, et~al.]{speech-recog}
Hannun, A., Case, C., Casper, J., Catanzaro, B., Diamos, G., Elsen, E.,
  Prenger, R., Satheesh, S., Sengupta, S., Coates, A., et~al.
\newblock Deep speech: Scaling up end-to-end speech recognition.
\newblock \emph{arXiv preprint arXiv:1412.5567}, 2014.

\bibitem[Hendrycks \& Gimpel(2016)Hendrycks and Gimpel]{baseline}
Hendrycks, D. and Gimpel, K.
\newblock A baseline for detecting misclassified and out-of-distribution
  examples in neural networks.
\newblock \emph{arXiv preprint arXiv:1610.02136}, 2016.

\bibitem[Hendrycks et~al.(2019{\natexlab{a}})Hendrycks, Mazeika, and
  Dietterich]{oe}
Hendrycks, D., Mazeika, M., and Dietterich, T.
\newblock Deep anomaly detection with outlier exposure.
\newblock \emph{In International Conference on Learning Representations},
  2019{\natexlab{a}}.

\bibitem[Hendrycks et~al.(2019{\natexlab{b}})Hendrycks, Mazeika, Kadavath, and
  Song]{self-supervised}
Hendrycks, D., Mazeika, M., Kadavath, S., and Song, D.
\newblock Using self-supervised learning can improve model robustness and
  uncertainty.
\newblock In \emph{Advances in Neural Information Processing Systems}, pp.\
  15663--15674, 2019{\natexlab{b}}.

\bibitem[Hoffmann(2007)]{pca}
Hoffmann, H.
\newblock Kernel {PCA} for novelty detection.
\newblock \emph{Pattern recognition}, 40\penalty0 (3):\penalty0 863--874, 2007.

\bibitem[H{\"u}llermeier \& Waegeman(2019)H{\"u}llermeier and
  Waegeman]{ML-uncertainty-survey}
H{\"u}llermeier, E. and Waegeman, W.
\newblock Aleatoric and epistemic uncertainty in machine learning: A tutorial
  introduction.
\newblock \emph{arXiv preprint arXiv:1910.09457}, 2019.

\bibitem[Jacobs et~al.(1991)Jacobs, Jordan, Nowlan, and Hinton]{moe}
Jacobs, R.~A., Jordan, M.~I., Nowlan, S.~J., and Hinton, G.~E.
\newblock Adaptive mixtures of local experts.
\newblock \emph{Neural computation}, 3\penalty0 (1):\penalty0 79--87, 1991.

\bibitem[Krizhevsky et~al.(2009)Krizhevsky, Hinton, et~al.]{cifar10}
Krizhevsky, A., Hinton, G., et~al.
\newblock Learning multiple layers of features from tiny images.
\newblock 2009.

\bibitem[Lakshminarayanan et~al.(2016)Lakshminarayanan, Pritzel, and
  Blundell]{proper_scoring}
Lakshminarayanan, B., Pritzel, A., and Blundell, C.
\newblock Simple and scalable predictive uncertainty estimation using deep
  ensembles.
\newblock \emph{arXiv preprint arXiv:1612.01474}, 2016.

\bibitem[LeCun et~al.(1998)LeCun, Bottou, Bengio, and Haffner]{lenet}
LeCun, Y., Bottou, L., Bengio, Y., and Haffner, P.
\newblock Gradient-based learning applied to document recognition.
\newblock \emph{Proceedings of the IEEE}, 86\penalty0 (11):\penalty0
  2278--2324, 1998.

\bibitem[Lee et~al.(2018)Lee, Lee, Lee, and Shin]{mahalanobis}
Lee, K., Lee, K., Lee, H., and Shin, J.
\newblock A simple unified framework for detecting out-of-distribution samples
  and adversarial attacks.
\newblock In \emph{Advances in Neural Information Processing Systems}, pp.\
  7167--7177, 2018.

\bibitem[Liang et~al.(2017)Liang, Li, and Srikant]{odin}
Liang, S., Li, Y., and Srikant, R.
\newblock Enhancing the reliability of out-of-distribution image detection in
  neural networks.
\newblock \emph{arXiv preprint arXiv:1706.02690}, 2017.

\bibitem[Maaten \& Hinton(2008)Maaten and Hinton]{t-SNE}
Maaten, L. v.~d. and Hinton, G.
\newblock Visualizing data using t-sne.
\newblock \emph{Journal of machine learning research}, 9\penalty0
  (Nov):\penalty0 2579--2605, 2008.

\bibitem[Majumder et~al.(2017)Majumder, Poria, Gelbukh, and
  Cambria]{text-analysis}
Majumder, N., Poria, S., Gelbukh, A., and Cambria, E.
\newblock Deep learning-based document modeling for personality detection from
  text.
\newblock \emph{IEEE Intelligent Systems}, 32\penalty0 (2):\penalty0 74--79,
  2017.

\bibitem[Netzer et~al.(2011)Netzer, Wang, Coates, Bissacco, Wu, and Ng]{svhn}
Netzer, Y., Wang, T., Coates, A., Bissacco, A., Wu, B., and Ng, A.~Y.
\newblock Reading digits in natural images with unsupervised feature learning.
\newblock 2011.

\bibitem[Papernot \& McDaniel(2018)Papernot and McDaniel]{dknn}
Papernot, N. and McDaniel, P.
\newblock Deep k-nearest neighbors: Towards confident, interpretable and robust
  deep learning.
\newblock \emph{arXiv preprint arXiv:1803.04765}, 2018.

\bibitem[Ruff et~al.(2018)Ruff, Vandermeulen, Goernitz, Deecke, Siddiqui,
  Binder, M{\"u}ller, and Kloft]{deep-svdd}
Ruff, L., Vandermeulen, R., Goernitz, N., Deecke, L., Siddiqui, S.~A., Binder,
  A., M{\"u}ller, E., and Kloft, M.
\newblock Deep one-class classification.
\newblock In \emph{International conference on machine learning}, pp.\
  4393--4402. PMLR, 2018.

\bibitem[Sch{\"o}lkopf et~al.(1999)Sch{\"o}lkopf, Williamson, Smola,
  Shawe-Taylor, Platt, et~al.]{oc-svm}
Sch{\"o}lkopf, B., Williamson, R.~C., Smola, A.~J., Shawe-Taylor, J., Platt,
  J.~C., et~al.
\newblock Support vector method for novelty detection.
\newblock In \emph{NIPS}, volume~12, pp.\  582--588. Citeseer, 1999.

\bibitem[Steinhardt \& Liang(2016)Steinhardt and Liang]{kl_div_entropy}
Steinhardt, J. and Liang, P.
\newblock Unsupervised risk estimation using only conditional independence
  structure.
\newblock \emph{arXiv preprint arXiv:1606.05313}, 2016.

\bibitem[Tipping \& Bishop(1999)Tipping and Bishop]{prob_pca}
Tipping, M.~E. and Bishop, C.~M.
\newblock Probabilistic principal component analysis.
\newblock \emph{Journal of the Royal Statistical Society: Series B (Statistical
  Methodology)}, 61\penalty0 (3):\penalty0 611--622, 1999.

\bibitem[Xiao et~al.(2017)Xiao, Rasul, and Vollgraf]{fashion-mnist}
Xiao, H., Rasul, K., and Vollgraf, R.
\newblock Fashion-mnist: a novel image dataset for benchmarking machine
  learning algorithms.
\newblock \emph{arXiv preprint arXiv:1708.07747}, 2017.

\bibitem[Yu et~al.(2015)Yu, Seff, Zhang, Song, Funkhouser, and Xiao]{lsun}
Yu, F., Seff, A., Zhang, Y., Song, S., Funkhouser, T., and Xiao, J.
\newblock Lsun: Construction of a large-scale image dataset using deep learning
  with humans in the loop.
\newblock \emph{arXiv preprint arXiv:1506.03365}, 2015.

\end{thebibliography}
